\documentclass[pdflatex,sn-mathphys-num]{sn-jnl}

\usepackage{multirow}
\usepackage{makecell}
\usepackage{graphicx}%
\usepackage{multirow}%
\usepackage{amsmath,amssymb,amsfonts}%
\usepackage{amsthm}%
\usepackage{mathrsfs}%
\usepackage[title]{appendix}%
\usepackage{xcolor}%
\usepackage{float}
\usepackage{textcomp}%
\usepackage{manyfoot}%
\usepackage{booktabs}%
\usepackage{enumitem}
\usepackage{algorithm}%
\usepackage{algorithmicx}%
\usepackage{algpseudocode}%
\usepackage{listings}%
\usepackage{longtable}
\usepackage{makecell}
\usepackage{multirow}
\usepackage{placeins}

\theoremstyle{thmstyleone}%
\theoremstyle{thmstyletwo}%

\theoremstyle{thmstylethree}%

\begin{document}

\title[Article Title]{
Comparative Study of Anatomical and Learned Features in AI Models for Structural Brain MRI
}


\author[1]{\fnm{Boyang} \sur{Yu}}

\equalcont{These authors contributed equally to this work.}

\author[1]{\fnm{Miquel} \spfx{Lopez} \sur{Escoriza}}

\equalcont{These authors contributed equally to this work.}

\author[1,2]{\fnm{Long} \sur{Chen}}

\author[2]{\fnm{Arjun V.} \sur{Masurkar}}

\author[2]{\fnm{Narges} \sur{Razavian}}

\author*[1,3]{\fnm{Carlos} \sur{Fernandez-Granda}}\email{cfgranda@cims.nyu.edu}

\affil[1]{\orgdiv{Center for Data Science}, \orgname{New York University, NYU}, \orgaddress{\street{60 Fifth Ave}, \city{New York}, \postcode{10011}, \state{New York}, \country{United States}}}

\affil[2]{\orgdiv{Department of Neurology}, \orgname{NYU Grossman School of Medicine}, \orgaddress{\street{ 550 1st Ave}, \city{New York}, \postcode{10016}, \state{New York}, \country{United States}}}

\affil[3]{\orgdiv{Courant Institute of Mathematical Sciences}, \orgname{NYU}, \orgaddress{\street{ 251 Mercer St}, \city{New York}, \postcode{10012}, \state{New York}, \country{United States}}}




\abstract{In this work, we comprehensively evaluate  three popular feature-extraction paradigms in AI-based neuroimaging modeling: (1) computation of anatomical surfaces and volumes, (2) supervised learning with convolutional neural networks (CNNs), and (3) unsupervised pretraining of vision transformer (ViT) foundation models, followed by supervised finetuning. Our study is based on 18 publicly available datasets containing 3D structural T1-weighted MRI scans from approximately 80,000 participants across seven distinct clinical tasks. We observe that a linear model based on anatomical features matches the diagnostic performance of complex nonlinear features learned by sophisticated AI frameworks, including foundation models trained on thousands of scans. Conversely, CNNs and pretrained ViTs learn features that implicitly capture relevant anatomical information, bypassing the need for explicit feature extraction. Building upon these insights, we propose Anatomy Segmentation Pretraining (ASP), a novel method to incorporate anatomical information during foundation-model pretraining, which outperforms existing models in biological age estimation.}
\keywords{Deep learning, pretraining, neuroimaging, brain anatomy, segmentation}



\maketitle

\section*{Introduction  }\label{intro}

\begin{figure}[h]
\makebox[\textwidth][c]{%
    \includegraphics[width=1.2\textwidth]{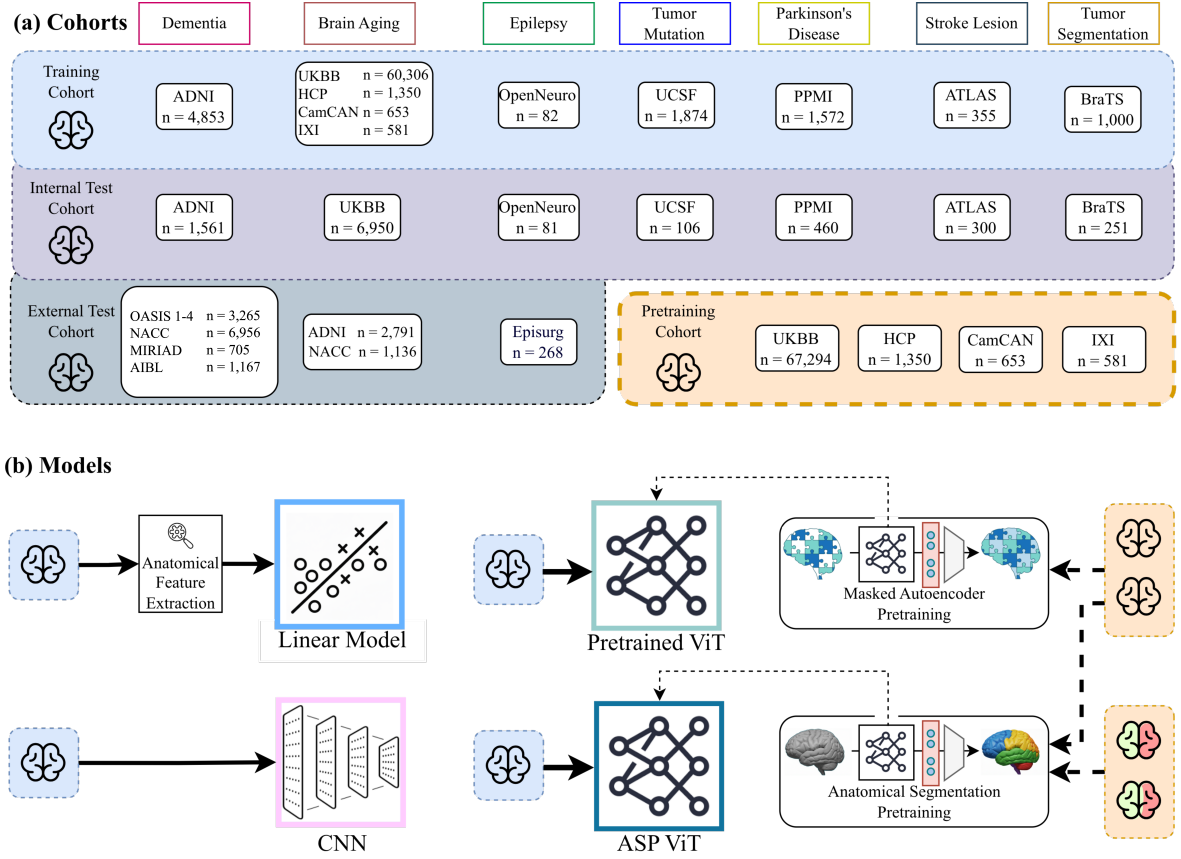}
}
\caption{\textbf{Overview of the study}. \textbf{(a) Cohorts used in the study.} Cohorts used for pretraining, downstream training, internal testing, and external evaluation. Reported $n$ values denote MRI scans; all splits were performed at the participant level, such that all scans from the same participant were assigned to the same split. External cohorts were drawn from independent data sources. \textbf{(b) AI models included in the study}. These represent the main feature-extraction approaches for neuroimaging data: linear models based on anatomical features, convolutional neural networks (CNNs) and vision transformers (ViTs) pretrained on large general-population cohort. For the ViTs, in addition to standard pretraining based on masked autoencoding, we propose a novel scheme called Anatomical Segmentation Pretraining (ASP).}
\label{fig:overview}
\end{figure}

Advances in machine learning and computer vision have exerted a major influence on neuroimaging over the past two decades. Initially, these models relied on hand-engineered features, such as hippocampal volume or cortical thickness, combined with classical statistical models (e.g. logistic regression or support vector machines)~\citep{kloppel2008automatic, wee2012identification,popuri2020using, hwang2022disentangling}. Subsequently, modern convolutional neural networks (CNNs) enabled automatic learning of complex features, which are challenging to interpret anatomically, but achieved strong performance for a variety of neuroimaging tasks ~\citep{payan2015predictingalzheimersdiseaseneuroimaging, basaia2019automated, chang2018deep, havaei2017brain}. More recently, foundation models based on vision transformers (ViTs) ~\citep{dosovitskiy2020image} are able to learn features from vast unlabeled datasets~\citep{caron2021emerging, he2022masked}. They can then be finetuned for different downstream tasks. Several recent works adapt this approach to neuroimaging data~\citep{hatamizadeh2022unetr, cox2024brainsegfounder, tak2024foundation,zhu20253d}, for example to develop multimodal models.

In this work, we perform a comparative study of AI models for neuroimaging, with a focus on 3D structural T1-weighted MRI scans. T1-weighted MRI is a standard imaging modality that enables visualization of anatomical detail at a high resolution and is a routine part of most clinical protocols. Figure~\ref{fig:overview} provides an overview of the study. We consider seven different clinical tasks: diagnostics of Alzheimer's disease, epilepsy, Parkinson's disease, identification of tumor IDH mutations, brain age prediction and segmentation of brain tumors and stroke lesions. Overall, we use 18 different publicly-available datasets with more than 96,000 MRI scans from approximately 80,000 participants for training and evaluation. 

Our overarching goal is to compare the three main strategies to obtain features for AI-based modeling in neuroimaging: (1) extraction of preselected anatomical characteristics such as volumes and surfaces of brain regions, (2) supervised learning with convolutional neural networks, and (3) unsupervised pretraining of foundation vision transformers on large general-population cohorts, followed by supervised finetuning. As reported in the Results section, the three strategies yield similar performance across the different tasks. Our results indicate that preselected anatomical features match the diagnostic performance of complex nonlinear features learned by sophisticated AI frameworks, including foundation models trained on thousands of scans. Conversely, convolutional neural networks and pretrained vision transformers learn features that implicitly capture relevant anatomical information, bypassing the need for computationally intensive feature extraction. 

In the case of ViTs pretrained on a large general-population cohort, we show that anatomical information is learned already during the pretraining stage, when the learned features become increasingly predictive of different brain structures. Inspired by this observation, we propose a novel pretraining strategy  (Anatomy Segmentation Pretraining or ASP), which aims to incorporate explicit anatomical grounding in the training of neuroimaging foundation models. As described in the Methods section, this is achieved by forcing the model to perform segmentation of relevant brain structures during pretraining. While this approach performs on par with the other strategies across the clinical tasks, it achieves superior performance in estimating biological age, a task for which a larger number of training data is available.

In summary, this comparative study demonstrates that explicit anatomical features remain competitive with representations learned by advanced deep learning models, particularly across clinical cohorts ranging from dozens to thousands of training examples. Furthermore, we introduce a novel, anatomy-guided pretraining framework that explicitly leverages anatomical information to enhance the pretraining of neuroimaging foundation models.

\begin{table}[t]
\caption{Datasets used to build the pretraining cohort, and the training and evaluation cohorts for each task in the study.}
\label{tab:datasets}
\centering
\begin{tabular}{l|c|c|c|c|c}
\hline
\textbf{Dataset} & \makecell{Number of\\ scans} & \makecell{Number of\\ subjects} & \makecell{Age (years)\\ (mean ± std)}  & \makecell{Gender\\ (female \%)} & \makecell{Brain volume in mL\\ (mean ± std)} \\ \hline
UKBB & 67294 & 62665 & 64.93 ± 7.69 & 52.81\% & 1156 ± 114 \\ 
HCP & 1350 & 1350 & 38.66 ± 25.76 & 54.97\% & 1179 ± 137 \\ 
CamCAN & 653 & 653 & 54.2 ± 18.5  & 49.35\% & 1156 ± 128 \\ 
IXI & 581 & 581 & 48.65 ± 16.47 & 53.87\% & 1137 ± 131 \\ \hline
NACC & 6956 & 4926 & 71.41 ± 11.25 & 53.34\% & 1034 ± 185 \\ 
ADNI & 6471 & 1817 & 74.68 ± 7.64 & 47.45\% & 1072 ± 116 \\ 
AIBL & 1167 & 664 & 74.56 ± 6.79 & 52.87\% & 1078 ± 110 \\ 
OASIS1 & 436 & 416 & 51.35 ± 25.27 & 61.47\% & 1089 ± 153 \\ 
OASIS2 & 373 & 150 & 77.01 ± 7.64 & 57.10\% & 1045 ± 124 \\ 
OASIS3 & 2275 & 1044 & 70.69 ± 9.18 & 48.39\% & 1061 ± 113 \\ 
OASIS4 & 181 & 180 & 71.09 ± 9.60 & 46.96\% & 1060 ± 116 \\ 
MIRIAD & 705 & 69 & 69.65 ± 6.88 & 55.51\% & 1039 ± 115 \\ 
PPMI & 2032 & 1306 & 63.29 ± 9.58 & 40.59\% & 1119 ± 157 \\ 
OpenNeuro & 163 & 163 & 6.67 ± 2.47 & 46.01\% & 1195 ± 144 \\ 
EPISURG & 268 & 268 & NA & NA & 1102 ± 133 \\ 
UCSF-PDGM & 1980 & 501 & 56.87 ± 15.02 & 40.32\% & NA \\ 
BraTS & 2502 & 1251 & NA & NA & 1103 ± 136 \\ 
ATLAS & 655 & 33 & NA & NA & 1445 ± 124 \\ \hline
\end{tabular}
\end{table}

\begin{figure}[t]%
\centering
\includegraphics[width=0.95\linewidth,height=0.78\textheight,keepaspectratio]{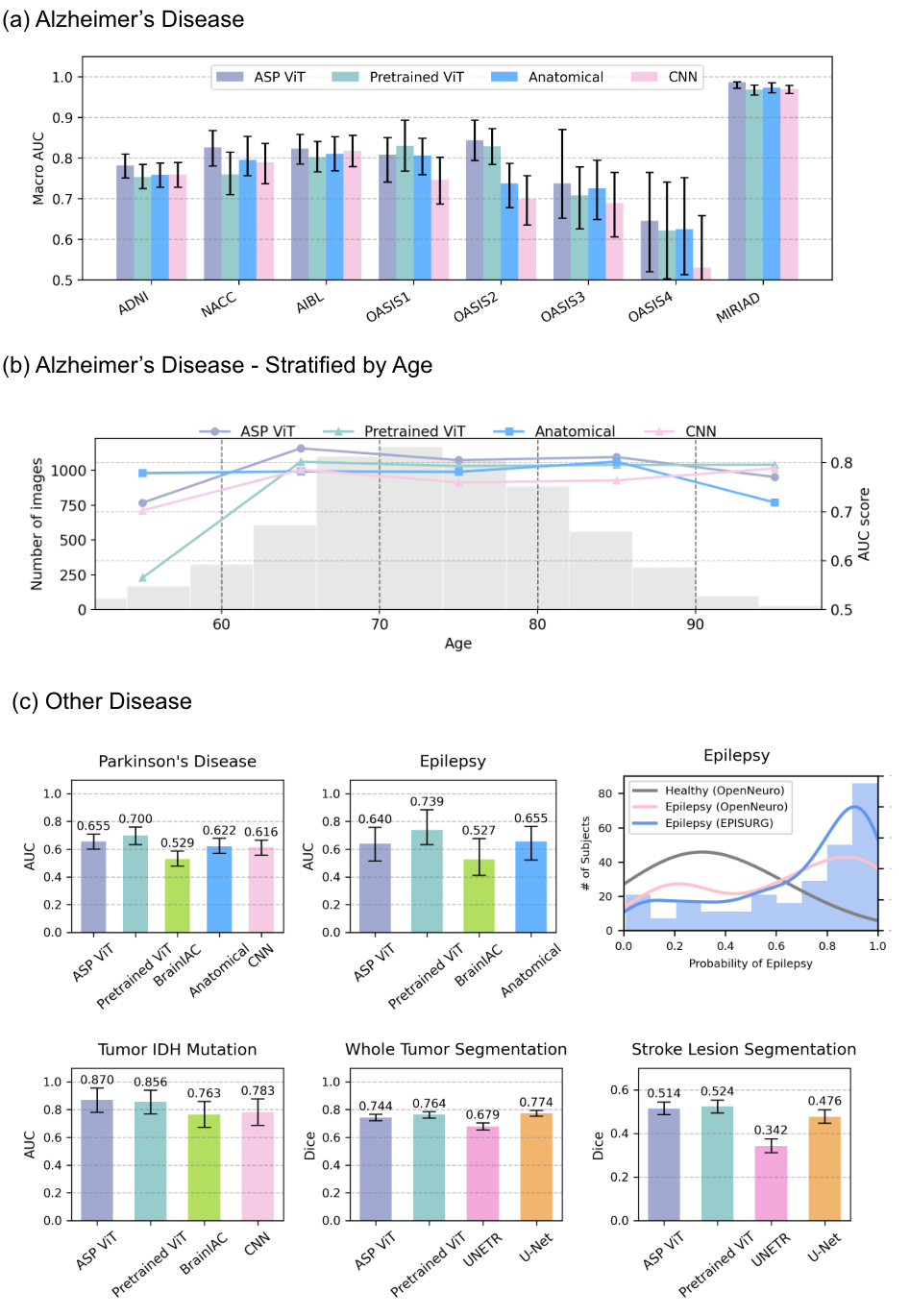}
\caption{\textbf{Performance comparison across diagnostic and segmentation tasks.} Alzheimer’s disease diagnosis was evaluated across an internal ADNI test cohort and seven external cohorts (NACC, AIBL, MIRIAD, and OASIS 1-4), where ASP ViT achieved the highest macro-AUC in most cohorts, though statistical significance over the CNN and anatomical logistic regression was only reached for OASIS 2. In the diagnostics of Parkinson’s disease (PPMI) and Epilepsy (OpenNeuro), the Pretrained ViT reached the highest AUC, showing a statistically significant improvement only over BrainIAC. Estimated probabilities for Epilepsy from ASP ViT are also shown for EPISURG, an additional evaluation set with only epilepsy cases. For the identification of brain tumor IDH mutations (UCSF-PDGM) ASP ViT yielded the highest AUC, though these differences were not statistically significant. CNN could not be applied to the epilepsy task due to limited training data. In segmentation tasks, a U-Net achieved the highest Dice score for adult brain gliomas (BraTS 2023), while the Pretrained ViT performed best for stroke lesion segmentation (ATLAS R2.0); in both cases, the differences were only statistically significant relative to UNETR. Overall, no significant performance differences were observed between the pretrained ViTs, CNNs, and anatomical logistic regression models across the evaluated tasks.
} \label{fig:disease-results}
\end{figure}

\section*{Results}\label{result}

Figure~\ref{fig:overview} provides an overview of the study. We compare the following AI models, adapted to process 3D MRI scans, across different tasks. The models are chosen to represent the main current approaches for image classification and segmentation in data-driven computer vision.   

\begin{itemize}
\item Linear logistic-regression models based on anatomical features consisting of 178 volumes and surfaces of brain regions (denoted Anatomical in the figures).
\item Convolutional neural networks, which are either ResNets~\citep{he2015deepresiduallearningimage} for classification tasks (denoted CNN) or U-Nets~\citep{ronneberger2015u} for segmentation tasks.
\item Vision-transformer models~\citep{dosovitskiy2020image} pretrained on a large pretraining cohort via masked autoencoding (denoted Pretrained ViT).
\item ViTs pretrained using a novel strategy (Anatomy Segmentation Pretraining, denoted ASP ViT) that leverages anatomical information extracted via segmentation from the scans in the pretraining cohort.
\end{itemize}

The pretraining cohort includes 69,878 MRI scans from 65,249 subjects, primarily from UK Biobank\citep{sudlow2015uk}. More details about the pretraining data are provided in the appendix (Section \ref{sec:pretraining_data}). In addition, we also include comparisons with BrainIAC~\citep{tak2024foundation}, a ViT pretrained using contrastive learning on a different pretraining cohort of 32,000 MRI scans, and UNETR~\citep{hatamizadeh2022unetr}, a medical image segmentation architecture that combines a ViT encoder with a U-Net–style decoder.     

The models are trained to perform the following seven tasks. The results are shown in Figure~\ref{fig:disease-results}. Additional details about the datasets used to build the training and evaluation cohorts for each task, and the pretraining cohort, are provided in Table~\ref{tab:datasets} and in the appendix (Section \ref{sec:datasets}). 

\begin{itemize}
\item \textbf{Diagnostics of Alzheimer's disease:} The goal is to distinguish between healthy controls, patients with mild cognitive impairment and patients with Alzheimer's disease.
The training cohort consists of n=1,453 subjects from ADNI~\citep{petersen2010alzheimer}. Evaluation was carried out on an internal test cohort extracted from ADNI (n=364), and seven external cohorts, extracted from NACC~\citep{beekly2007national} (n=4,926), AIBL~\citep{ellis2006australian} (n=664), MIRIAD~\citep{malone2013miriad} (n=69), and OASIS-1 (n=416), OASIS-2 (n=150), OASIS-3 (n=1,044) and OASIS-4 (n=180)~\citep{lamontagne2019oasis,marcus2010open, marcus2007open}.
For the internal test cohort and six external cohorts, ASP ViT has the highest macro-AUC, but the difference with respect to the CNN and the anatomical logistic regression model is only statistically significant for OASIS 2. We didn't apply BrainIAC to this task because its pretraining cohort includes several of our external test cohorts.
\item \textbf{Diagnostics of Parkinson's disease:} The goal is to distinguish between patients with Parkinson's and healthy controls. A training cohort (n=1,572) and an internal test cohort (n=460) were extracted from PPMI~\citep{marek2011parkinson}. The Pretrained ViT achieves the highest AUC, but the difference is only statistically significant with respect to BrainIAC. 
\item \textbf{Detection of tumor IDH mutations:} The goal is to identify brain tumors with Isocitrate Dehydrogenase (IDH) mutation. A training cohort (n=1,874) and an internal test cohort (n=106) were extracted from UCSF-PDGM~\citep{calabrese2022university}, which contains labels obtained via genetic sequencing of tissue acquired during biopsy or resection. ASP ViT achieves the highest AUC, but the difference with the remaining models is not statistically significant. 
\item \textbf{Diagnostics of epilepsy:} The goal is to distinguish between patients with epilepsy and healthy controls. A training cohort (n=82) and an internal test cohort (n=81) were extracted from OpenNeuro~\citep{schuch2023open}. The models were also applied to Episurg  \citep{perez2021self}, an independent cohort containing only epilepsy cases (n=268). 
The Pretrained ViT achieves the highest AUC, but the difference with the remaining models is not statistically significant. We could not apply the CNN to this task due to the limited training data.
\item \textbf{Whole tumor segmentation:} The goal is delineation of adult brain gliomas. A training cohort (n=1,000) and an internal test cohort (n=251) were extracted from the BraTS 2023 Adult Glioma challenge~\citep{baid2021rsna, menze2014multimodal, bakas2017advancing}. The U-Net achieved the highest Dice score (slightly above ASP ViT), but the difference is only statistically significant with respect to UNETR. If the training data is reduced to 10\%, ASP and Pretrained ViT outperform U-Net and UNETR.
 \item \textbf{Segmentation of stroke lesions:} The goal is segmentation of brain lesions associated with stroke. A training cohort (n=355) and an internal test cohort (n=300) were extracted from ATLAS R2.0~\citep{liew2022large}. The Pretrained ViT achieved the highest Dice score (slightly above ASP ViT), but the difference is only statistically significant with respect to UNETR.
\end{itemize}

Overall, we observed that there was no significant difference between the pretrained ViTs, the CNNs and linear models based on anatomical features across the different tasks. 

We conjecture that the inability of the more complex deep-learning models to outperform the simpler linear model is due to the limited training data available for each task. In order to investigate this further, we trained the models to estimate biological age using a much larger training cohort drawn from UK Biobank, HCP, CamCAN and IXI (n=62,890). Evaluation was carried out on a held-out subcohort from UK Biobank (n=6,950), and on two external cohorts of healthy subjects extracted from ADNI (n=2,791) and NACC (n=1,136).

\begin{figure}[tp!]%
\centering
  \includegraphics[width=0.95\linewidth,height=0.79\textheight,keepaspectratio]{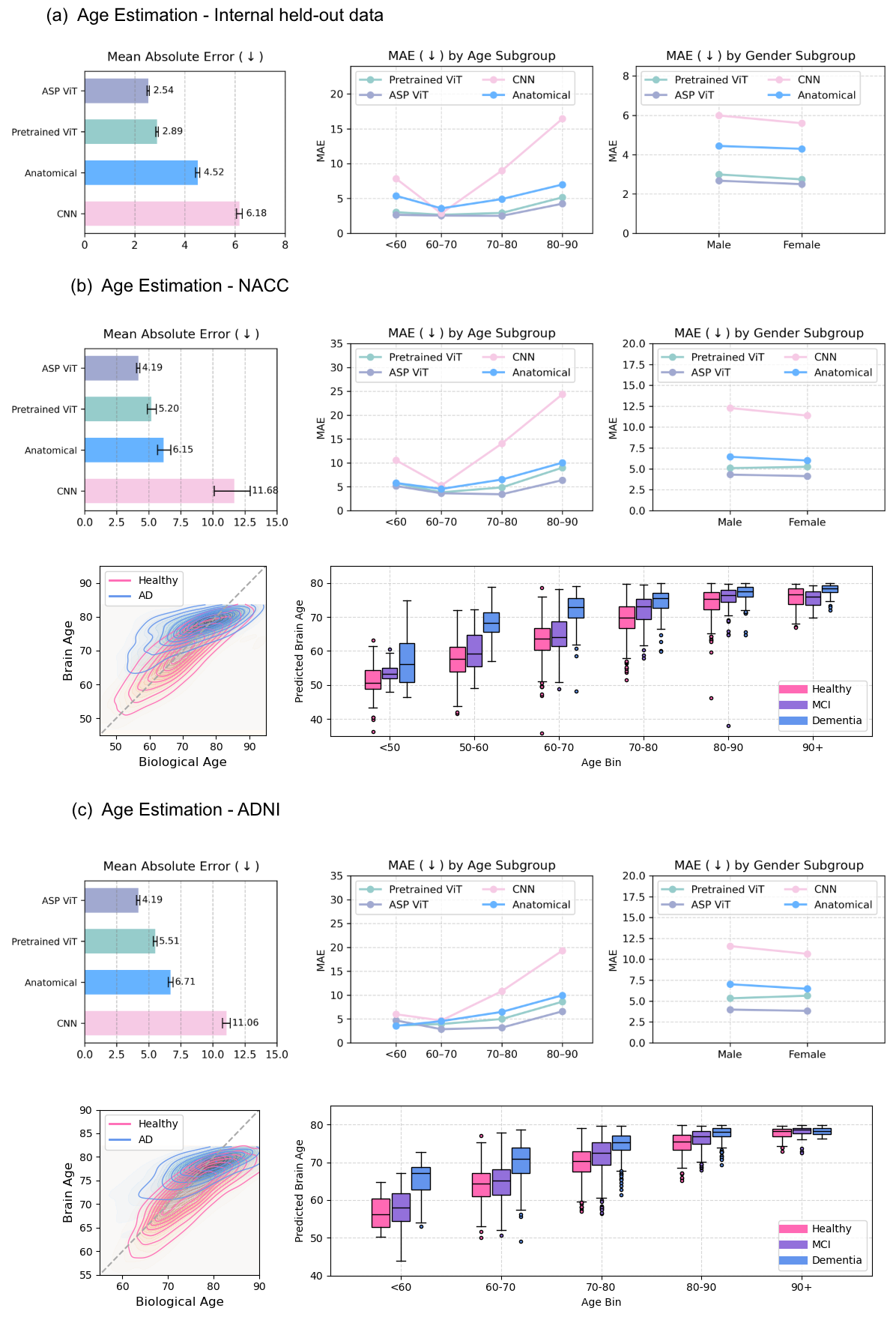}
\caption{\textbf{Age estimation performance across internal and external cohorts.} The ASP ViT model demonstrates superior performance, achieving a mean absolute error (MAE) of 2.54 years in the internal test cohort (a) and 4.19 years in both the NACC (b) and ADNI (c) external test cohorts. While the Pretrained ViT maintains competitive performance on the internal cohort with an MAE only 0.35 years greater than the ASP ViT, this performance gap widens to over 1 year in the external cohorts. Logistic regression underperforms relative to the ViT-based approaches by a margin of 1 to 2 years, though it consistently outperforms the CNN. These performance comparisons remain stable when the data is stratified by sex and age. Age-based stratification reveals more pronounced errors for individuals over 80, a demographic less represented in the training and pretraining datasets. When the best-performing model (ASP ViT) is applied to clinical populations, age estimates for dementia patients from the NACC (b) and ADNI (c) datasets are consistently higher than their biological age across most age intervals.
} \label{fig:aging}
\end{figure}

Figure~\ref{fig:aging} shows the results on age estimation. The ASP ViT model outperforms the rest, achieving a mean absolute error (MAE) of 2.54 years in the internal test cohort (a) and of 4.19 years in the NACC test cohort (b) and the ADNI test cohort (c). The Pretrained ViT is close in performance on the internal test cohort (0.35 years greater), but the gap widens for the external cohorts (more than 1 year greater). The logistic regression model performs worse than the ViT approaches by a significant margin (between 1 and 2 years), but clearly outperforms the CNN. The performance differences are maintained when stratifying the data by sex. Stratifying by age reveals that the differences between the models are mainly driven by the superior performance of the ViT models (particularly ASP ViT) on subjects with advanced ages (above 80), which are less represented in the training and pretraining cohorts, especially with respect to the CNN. 

The best-performing age-estimation model was also applied to mildly-cognitively impaired (MCI) and dementia patients from the NACC and ADNI  datasets, as shown in Figure~\ref{fig:aging}(b,c). Age estimates for dementia patients were consistently above their biological age, in most age intervals for both cohorts. The effect also exists to some extent for MCI subjects, but is less pronounced.

\section*{Methods}\label{sec:method}

In this section, we first present a new technique for pretraining vision transformers (ViTs) using anatomical information. Then, we describe the remaining methodology in our study, including masked autoencoder pretraining of ViTs, downstream application of pretrained ViTs, linear models based on anatomical features, convolutional neural networks, and data preprocessing.

\subsection*{Anatomy Segmentation Pretraining of Vision Transformer}

We propose a novel approach to pretrain ViTs on MRI data. The goal is to inject anatomical information into the model during pretraining. The motivation is that simple models that rely only on anatomical information, such as the logistic-regression model with brain-region volume and surface features used in this study, can be very effective (see Results section). 

\begin{figure}[h]
  \centering
  \includegraphics[width=1\linewidth]{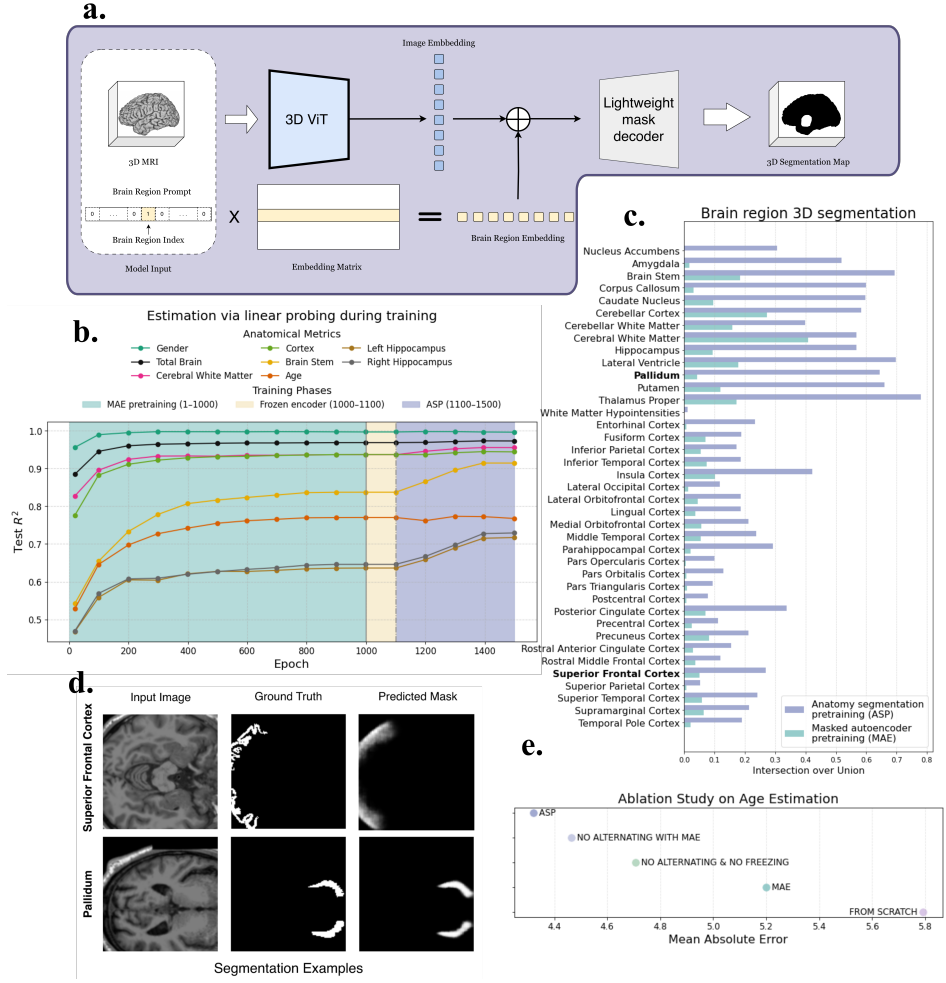}
\caption{\textbf{a. Prompt-based segmentation architecture}. The input consists of an image and a brain region index, which is used to retrieve the corresponding brain region embedding. This embedding is added to the image representation, and the combined features are passed through a lightweight decoder to predict the segmentation mask for the specified region. \textbf{b. Linear‑probe accuracy/$R^{2}$} on six target variables—two global attributes (age, gender) and four region‑specific volumetric measures—tracked over pre‑training epochs. The differences in overall accuracy across regions arise because some regions are inherently easier to predict than others. For example, cortex volume tends to be more accurately segmented due to its larger size and higher contrast, compared to smaller and less distinct regions like the hippocampus. For demographic statistics the same happens. Gender is easier to infer than age since it is a binary attribute, while age requires finer-grained regression. \textbf{c–d. Validation IoU by Brain Region.} Regional performance varies based on anatomical complexity; for instance, the superior frontal cortex's small, irregular geometry results in lower IoU, whereas the smoother, more consistent morphology of the pallidum allows for higher segmentation accuracy. \textbf{e. ASP training ablation.} Age-estimation performance for different ASP training variants on the external NACC cohort. The full ASP model achieves the lowest error, while removing alternating MAE updates, removing encoder freezing, using MAE alone, or training from scratch reduces performance. Detailed results are provided in Appendix Section~\ref{ablation_study}.}
  \label{fig:ASP}
\end{figure}

We build upon a key observation: when pretraining ViTs via masked autoencoding (MAE), the model automatically learns embeddings that reflect anatomical information, but only to some extent. The green region in the graph from Figure \ref{fig:ASP}(b) shows the accuracy of predicting the volumes of different brain regions from MAE embeddings via linear probing during training. The accuracy increases, indicating that the embeddings gradually capture anatomical information. In order to further boost the anatomical information learned by the model, we propose to incorporate brain-region segmentation as an auxiliary task during pretraining. 

Our procedure to perform anatomy segmentation pretraining (ASP) is illustrated in Figure \ref{fig:ASP}(a). A 3D ViT produces the embedding used for downstream tasks. After MAE pretraining (green region in Figure \ref{fig:ASP}(b)), we train a lightweight mask decoder to perform segmentation of different brain regions in two phases. First, the ViT parameters are frozen to ensure that the decoder learns how to use the existing embedding for segmentation (yellow region in Figure \ref{fig:ASP}(b)). Next, the encoder and decoder are trained jointly. In this ASP phase, segmentation and MAE pretraining are performed in alternating epochs to prevent the model from \textit{forgetting} useful features learned via MAE. This strategy is common in continual-learning methods that balance new-task adaptation with the retention of prior knowledge \cite{Kirkpatrick2017,li2017,rebuffi2017}.

Regions for ASP were selected based on their disease relevance and anatomical scale. We discarded regions that lacked a strong disease association and contributed less than 1\% of the total brain volume. To further improve efficiency, symmetrical structures (such as the left and right hippocampus) and anatomically related sub-units (such as segments of the corpus callosum) were consolidated into single labels. This resulted in 39 distinct regions, listed in Figure \ref{fig:ASP}(c). More details are provided in the Appendix Section \ref{sec:anatomical_segmentation_workflow}.

The segmentation architecture used to perform ASP is tailored specifically to the pretraining task. Deep-learning segmentation models typically rely heavily on skip connections, which emphasize local feature fusion and reduce the importance of global representations ~\cite{ronneberger2015u}, ~\cite{hatamizadeh2022unetr}. In contrast, the goal of ASP is to concentrate the anatomical information in the ViT embedding. To address this, we use a bottleneck decoder architecture(see Figure \ref{fig:seg_decoder}), where the segmentation prediction is generated directly from the ViT embedding, without any skip connections. The decoder is lightweight, in order to further encourage the model to concentrate rich anatomical information in the ViT transformer. More details are provided in the Appendix Section \ref{sec:seg_architecture}.

A key challenge is applying ASP to a relatively large number of anatomical regions (39, as explained above). A naive implementation based on a standard softmax output would be too memory intensive, as it would require generating 39 different 3D logit maps (39×128×128×128). Instead, we adopt a prompt-based architecture, inspired by \cite{liu2024uncertainty}, where a prompt is used to adapt the Segment Anything Model~\cite{kirillov2023segment} to different segmentation tasks. A \textit{brain-region prompt} indicates which brain region should be segmented by the lightweight decoder, and the decoder output is compared to a binary segmentation task indicating the appropriate region (see Figure \ref{fig:ASP}(a)). The brain region is selected at random during pretraining. 

We trained the ASP model for 400 epochs with a batch size of 16 per GPU. We used binary cross entropy and the AdamW optimizer ($\beta_{1} = 0.9$, $\beta_{2} = 0.95$, weight decay = 0.04), with a base learning rate of $5\times10^{-4}$. We applied learning rate scheduling via cosine annealing with linear warm-up over the first 5 epochs. Convergence is observed at around 200 epochs, requiring approximately one month of training on a single 42GB NVIDIA RTX8000 GPU.

Figure \ref{fig:ASP}(c) reports the performance of the decoder for the different regions before and after the ASP phase, showcasing that the model is indeed able to learn how to segment the different regions reasonably accurately (although its performance for smaller or more intricate region is relatively modest). Figure \ref{fig:ASP}(d) shows some examples of segmentation outputs, compared to the ground truth. Figure \ref{fig:ASP}(e) provides an ablation study, which establishes that each element in the proposed framework (prior MAE pretraining, initial freezing of the encoder to train the decoder, and alternating with MAE to avoid forgetting) all provide a non-negligible increase in performance for the downstream task of age estimation. Further information about the ablation study is provided in \ref{ablation_study}.

\subsection*{Masked Autoencoder Pretraining of ViTs}

Masked Autoencoding (MAE) is a self-supervised learning framework that trains ViTs by reconstructing masked images, which are missing a random subset of their pixels~\citep{he2022masked}. We adapted this framework to process 3D structural MRI volumes. We used a ViT-Base encoder adapted to 3D inputs, consisting of 12 transformer blocks with 12 attention heads and a 768-dimensional embedding, using 16×16×16-voxel patches. During pretraining, 75\% of the 3D patches in each volume were randomly masked, and the ViT was trained to reconstruct the original volumes from the unmasked patches. Volumes were represented as single-channel inputs with intensities scaled to $[0,1]$; images were padded to $256 \times 256 \times 256$, randomly cropped to $128 \times 128 \times 128$, and augmented with random flipping and intensity shifting. Pretraining used mean squared error (MSE) reconstruction loss for 1000 epochs with batch size 32 per GPU, AdamW~\citep{loshchilov2017decoupled} ($\beta_1=0.9$, $\beta_2=0.95$, weight decay $=0.005$), and a cosine learning rate schedule~\citep{loshchilov2016sgdr} with base learning rate 0.0015 and linear warm-up during the first 2.5\% of training steps.  Convergence is observed at around 900 epochs, which took around 10 days on two 80GB NVIDIA A100 GPUs.

\subsection*{Downstream Application of Pretrained ViTs}

To adapt the pretrained ViTs for downstream applications, we feed the resulting embeddings into task-specific output modules: a binary-classification head for diagnostic tasks, a linear-regression head for regression tasks (e.g. estimation of age or of anatomical features) and a U-Net–style decoder for voxel-level segmentation, following the UNETR architecture~\citep{hatamizadeh2022unetr}. See Section \ref{sec:mae_additional} for additional details. The pretrained ViTs were adapted to the different downstream tasks using the following strategies. 

\begin{itemize}
\item \textbf{Linear probing:} The linear regression and classification heads were trained to generate predictions from the ViT embedding, using the \texttt{sklearn} implementations of logistic regression and ordinary least squares. This approach was used as a diagnostic tool during pretraining to monitor representation quality using anatomical and demographic targets, as illustrated by Figure \ref{fig:ASP}(b). It was also used for diagnostics of Parkinson's disease and detection of tumor IDH Mutation, due to the small size of the training datasets.


\item \textbf{Low-Rank Adaptation (LoRA):} Both ViT encoders (ASP and pretrained) were adapted using LoRA ~\citep{zhu2024melo, hu2022lora}: the original model parameters were kept frozen, and trainable low-rank matrices were introduced into the attention layers. These low-rank parameters were optimized jointly with the task-specific heads, substantially reducing the number of trainable parameters compared with updating the entire encoder. Random cropping was employed for regularization during training. At inference time, we used center cropping for classification tasks and a sliding window approach for full-volume segmentation tasks. Training was performed with the AdamW optimizer for 150 epochs. 

\item \textbf{Full Finetuning:} For ablation experiments, we updated all parameters from the ASP ViT encoder jointly with the downstream head using the same optimization and augmentation settings as LoRA. See Appendix Section \ref{sec:additional_model_training_details}.

\end{itemize}

\subsection*{Linear Models with Anatomical Features}

For each of our downstream diagnostic tasks, as well as age estimation, we trained linear models based on 178 anatomical features extracted via FreeSurfer~\citep{fischl2012freesurfer} (v7.4.1), using the \texttt{recon-all} pipeline. The features include regional volume measurements, cortical surface areas, and cortical thickness values. A complete list is provided in Appendix Table~\ref{tab:anatomical_features}. The resulting models are referred to as \emph{Anatomical} in the Results section. These linear models achieved similar performance as nonlinear alternatives; details are provided in Appendix Section~\ref{sec:lr_vs_rf}. For Alzheimer's diagnosis, we encoded healthy as 0, Alzheimer's as 1 and mild cognitive impairment as 0.5, as this produced better results than treating diagnosis as a multiclass classification task~\citep{liu2022generalizable}, see Appendix Section~\ref{sec:label_encoding}. 

\subsection*{Convolutional Neural Networks}
We used the ResNet-18 architecture~\citep{he2015deepresiduallearningimage} for diagnostic classification tasks and the U-Net architecture~\citep{ronneberger2015u} for segmentation tasks, based on preliminary experiments. For classification, the models were trained for up to 60 epochs using a batch size of 16, stochastic gradient descent (SGD)~\citep{ruder2017overviewgradientdescentoptimization}, an initial learning rate of 0.01, and early stopping based on validation loss. For Alzheimer's diagnosis, we used the same label encoding as in the logistic regression model, and optimized a binary cross-entropy objective during training. Additional analyses on baseline CNN selection are reported in Appendix Sections~\ref{sec:preliminary_analysis}.

\subsection*{Data Preprocessing}
To harmonize data across cohorts, we implemented a standardized preprocessing pipeline for all MRIs. While datasets for segmentation tasks (BraTS and ATLAS R2.0) were already standardized, all other raw data—initially in DICOM or NIFTI format—underwent a uniform processing sequence. DICOM series were first converted to NIFTI using the dcm2nii tool.

Except for the pre-standardized segmentation cohorts, scans were processed using FreeSurfer’s recon-all workflow, which consists of three sequential phases (initial processing, surface reconstruction and refinement). This workflow also served as our primary quality control mechanism; any images that failed the initial processing phase were excluded from the study. A detailed description of the recon-all execution time and workflow is provided in Appendix Section \ref{sec_app:image_data_harmonization}.

Finally, to ensure spatial alignment, all images and tissue masks were coregistered to the MNI152 2009a Nonlinear Symmetric template using the mri\_vol2vol command. The resulting preprocessed scans have a standardized volume of 197 × 233 × 189 voxels with an isotropic resolution of one cubic millimeter. 

\subsection*{Statistical analysis}
In each experiment, we report the overall metric and 95\% confidence interval (CI), calculated by bootstrapping the held-out test set 1000 times. Specifically, we generate 1000 bootstrap samples by resampling the test set with replacement, compute the metric on each sample, and estimate the 95\% CI from the 2.5th and 97.5th percentiles of the resulting metric distribution.

\subsection*{Ethics Approval and Consent to Participate}
This study used de-identified, secondary data from publicly available neuroimaging repositories (UKBB, HCP, CamCAN, IXI, NACC, ADNI, OASIS, AIBL, MIRIAD, PPMI, OpenNeuro, EPISURG, UCSF-PDGM, BraTS, ATLAS). Each dataset was collected under institutional ethics approval and informed consent obtained by the original data-collecting institutions, as described in their respective publications. No new human data were collected for this study.

\section*{Discussion}\label{discussion}

Our study demonstrates that across a wide range of publicly available structural MRI datasets, linear models trained on preselected anatomical features remain highly competitive with deep neural networks. While traditional segmentation approaches are often computationally expensive for feature extraction, modern deep learning-based segmentation frameworks (e.g., FastSurfer \citep{henschel2020fastsurfer}) offer an efficient alternative. Our results suggest that leveraging these deep-learning tools to extract structured anatomical features for linear models provides a computationally efficient, interpretable, and robust baseline for various diagnostic tasks.

Furthermore, our results indicate that diagnostic performance using structural MRI may have reached a saturation point within existing public datasets that contain only hundreds to a few thousand scans. Further progress will likely require the curation of substantially larger cohorts. Our age-estimation experiments on a cohort of 60,000 subjects indicate that scaling public datasets could unlock further methodological innovation. 

From the methodological perspective, we show that vision transformers learn relevant anatomical structure during unsupervised pretraining, but at a slow rate that can be greatly accelerated using anatomical segmentation as an auxiliary pretraining task. This suggests that leveraging clinically-relevant tasks could mitigate the small size of medical pretraining datasets when compared to general computer-vision pretraining databases with millions of examples. 

\section*{Data availability}
Data from UK Biobank, ADNI, NACC, AIBL, OASIS 1--4, MIRIAD, PPMI, HCP, CamCAN, IXI, OpenNeuro, EPISURG, UCSF-PDGM, BraTS, and ATLAS are available to qualified researchers through application to the respective consortia (see Appendix A for links and access details). No new data were generated for this study.

\section*{Code availability}
The code used to train and evaluate the models described in this study is publicly available at \url{https://github.com/MIQUEL-LOPEZ-ESCORIZA/ai4brainmri}.

\section*{Acknowledgements}
This work was supported by Alzheimer’s Association grant AARG-NTF-21-848627 and NSF grant 2404476, as well as by the NYU IT High Performance Computing. The funders played no role in study design, data collection, analysis, interpretation, or writing of this manuscript.  Dataset-specific acknowledgements are provided in Appendix Section \ref{sec_app:acknowledgement}.

\section*{Author contributions}
B.Y. and M.L.E. designed and implemented the models, ran experiments, analyzed results. L.C. conducted the LoRa experiments. A.V.M. and N.R. provided clinical guidance and interpretation. C.F-G. conceived and supervised the study. All authors contributed to writing and approved the final manuscript.

\section*{Competing interests}
The authors declare no competing interests.

\bibliography{sn-bibliography}

\backmatter

\section*{Supplementary information}





\begin{appendices}

\section{Dataset Information}\label{sec:datasets}
\subsection{Acknowledgement}
\label{sec_app:acknowledgement}
This research has been conducted using the UK Biobank Resource under Application Number 166231.
Research reported in this publication was supported by the National Institute On Aging of the National Institutes of Health under Award Number U01AG052564 and by funds provided by the McDonnell Center for Systems Neuroscience at Washington University in St. Louis. The HCP-Aging 2.0 Release data used in this report came from DOI: 10.15154/1520707

Data collection and sharing for this project was provided by the Cambridge Centre for Ageing and Neuroscience (CamCAN). CamCAN funding was provided by the UK Biotechnology and Biological Sciences Research Council (grant number BB/H008217/1), together with support from the UK Medical Research Council and University of Cambridge, UK.
We acknowledge the use of data from the IXI dataset (https://brain-development.org/ixi-dataset/), which was made available through the Imaging in eXcellence Initiative (EPSRC GR/S21533/02).

The NACC database is funded by NIA/NIH Grant U24 AG072122. SCAN is a multi-institutional project that was funded as a U24 grant (AG067418) by the National Institute on Aging in May 2020. Data collected by SCAN and shared by NACC are contributed by the NIA-funded ADRCs as follows:

Arizona Alzheimer’s Center - P30 AG072980 (PI: Eric Reiman, MD); R01 AG069453 (PI: Eric Reiman (contact), MD); P30 AG019610 (PI: Eric Reiman, MD); and the State of Arizona which provided additional funding supporting our center; Boston University - P30 AG013846 (PI Neil Kowall MD); Cleveland ADRC - P30 AG062428 (James Leverenz, MD); Cleveland Clinic, Las Vegas – P20AG068053; Columbia - P50 AG008702 (PI Scott Small MD); Duke/UNC ADRC – P30 AG072958; Emory University - P30AG066511 (PI Levey Allan, MD, PhD); Indiana University - R01 AG19771 (PI Andrew Saykin, PsyD); P30 AG10133 (PI Andrew Saykin, PsyD); P30 AG072976 (PI Andrew Saykin, PsyD); R01 AG061788 (PI Shannon Risacher, PhD); R01 AG053993 (PI Yu-Chien Wu, MD, PhD); U01 AG057195 (PI Liana Apostolova, MD); U19 AG063911 (PI Bradley Boeve, MD); and the Indiana University Department of Radiology and Imaging Sciences; Johns Hopkins - P30 AG066507 (PI Marilyn Albert, Phd.); Mayo Clinic - P50 AG016574 (PI Ronald Petersen MD PhD); Mount Sinai - P30 AG066514 (PI Mary Sano, PhD); R01 AG054110 (PI Trey Hedden, PhD); R01 AG053509 (PI Trey Hedden, PhD); New York University - P30AG066512-01S2 (PI Thomas Wisniewski, MD); R01AG056031 (PI Ricardo Osorio, MD); R01AG056531 (PIs Ricardo Osorio, MD; Girardin Jean-Louis, PhD); Northwestern University - P30 AG013854 (PI Robert Vassar PhD); R01 AG045571 (PI Emily Rogalski, PhD); R56 AG045571, (PI Emily Rogalski, PhD); R01 AG067781, (PI Emily Rogalski, PhD); U19 AG073153, (PI Emily Rogalski, PhD); R01 DC008552, (M.-Marsel Mesulam, MD); R01 AG077444, (PIs M.-Marsel Mesulam, MD, Emily Rogalski, PhD); R01 NS075075 (PI Emily Rogalski, PhD); R01 AG056258 (PI Emily Rogalski, PhD); Oregon Health and Science University - P30 AG008017 (PI Jeffrey Kaye MD); R56 AG074321 (PI Jeffrey Kaye, MD); Rush University - P30 AG010161 (PI David Bennett MD); Stanford – P30AG066515; P50 AG047366 (PI Victor Henderson MD MS); University of Alabama, Birmingham – P20; University of California, Davis - P30 AG10129 (PI Charles DeCarli, MD); P30 AG072972 (PI Charles DeCarli, MD); University of California, Irvine - P50 AG016573 (PI Frank LaFerla PhD); University of California, San Diego - P30AG062429 (PI James Brewer, MD, PhD); University of California, San Francisco - P30 AG062422 (Rabinovici, Gil D., MD); University of Kansas - P30 AG035982 (Russell Swerdlow, MD); University of Kentucky - P30 AG028283-15S1 (PIs Linda Van Eldik, PhD and Brian Gold, PhD); University of Michigan ADRC - P30AG053760 (PI Henry Paulson, MD, PhD) P30AG072931 (PI Henry Paulson, MD, PhD) Cure Alzheimer's Fund 200775 - (PI Henry Paulson, MD, PhD) U19 NS120384 (PI Charles DeCarli, MD, University of Michigan Site PI Henry Paulson, MD, PhD) R01 AG068338 (MPI Bruno Giordani, PhD, Carol Persad, PhD, Yi Murphey, PhD) S10OD026738-01 (PI Douglas Noll, PhD) R01 AG058724 (PI Benjamin Hampstead, PhD) R35 AG072262 (PI Benjamin Hampstead, PhD) W81XWH2110743 (PI Benjamin Hampstead, PhD) R01 AG073235 (PI Nancy Chiaravalloti, University of Michigan Site PI Benjamin Hampstead, PhD) 1I01RX001534 (PI Benjamin Hampstead, PhD) IRX001381 (PI Benjamin Hampstead, PhD); University of New Mexico - P20 AG068077 (Gary Rosenberg, MD); University of Pennsylvania - State of PA project 2019NF4100087335 (PI David Wolk, MD); Rooney Family Research Fund (PI David Wolk, MD); R01 AG055005 (PI David Wolk, MD); University of Pittsburgh - P50 AG005133 (PI Oscar Lopez MD); University of Southern California - P50 AG005142 (PI Helena Chui MD); University of Washington - P50 AG005136 (PI Thomas Grabowski MD); University of Wisconsin - P50 AG033514 (PI Sanjay Asthana MD FRCP); Vanderbilt University – P20 AG068082; Wake Forest - P30AG072947 (PI Suzanne Craft, PhD); Washington University, St. Louis - P01 AG03991 (PI John Morris MD); P01 AG026276 (PI John Morris MD); P20 MH071616 (PI Dan Marcus); P30 AG066444 (PI John Morris MD); P30 NS098577 (PI Dan Marcus); R01 AG021910 (PI Randy Buckner); R01 AG043434 (PI Catherine Roe); R01 EB009352 (PI Dan Marcus); UL1 TR000448 (PI Brad Evanoff); U24 RR021382 (PI Bruce Rosen); Avid Radiopharmaceuticals / Eli Lilly; Yale - P50 AG047270 (PI Stephen Strittmatter MD PhD); R01AG052560 (MPI: Christopher van Dyck, MD; Richard Carson, PhD); R01AG062276 (PI: Christopher van Dyck, MD); 1Florida - P30AG066506-03 (PI Glenn Smith, PhD); P50 AG047266 (PI Todd Golde MD PhD)

Data collection and sharing for this project was funded by the Alzheimer's Disease Neuroimaging Initiative (ADNI) (National Institutes of Health Grant U01 AG024904) and DOD ADNI (Department of Defense award number W81XWH-12-2-0012). ADNI is funded by the National Institute on Aging, the National Institute of Biomedical Imaging and Bioengineering, and through generous contributions from the following: AbbVie, Alzheimer’s Association; Alzheimer’s Drug Discovery Foundation; Araclon Biotech; BioClinica, Inc.; Biogen; Bristol-Myers Squibb Company; CereSpir, Inc.; Cogstate; Eisai Inc.; Elan Pharmaceuticals, Inc.; Eli Lilly and Company; EuroImmun; F. Hoffmann-La Roche Ltd and its affiliated company Genentech, Inc.; Fujirebio; GE Healthcare; IXICO Ltd.; Janssen Alzheimer Immunotherapy Research \& Development, LLC.; Johnson \& Johnson Pharmaceutical Research \& Development LLC.; Lumosity; Lundbeck; Merck \& Co., Inc.; Meso Scale Diagnostics, LLC.;  NeuroRx Research; Neurotrack Technologies; Novartis Pharmaceuticals Corporation; Pfizer Inc.; Piramal Imaging; Servier; Takeda Pharmaceutical Company; and Transition Therapeutics. The Canadian Institutes of Health Research is providing funds to support ADNI clinical sites in Canada. Private sector contributions are facilitated by the Foundation for the National Institutes of Health (www.fnih.org). The grantee organization is the Northern California Institute for Research and Education, and the study is coordinated by the Alzheimer’s Therapeutic Research Institute at the University of Southern California. ADNI data are disseminated by the Laboratory for Neuro Imaging at the University of Southern California.

Data were provided in part by OASIS. Specifically:
OASIS-1: Cross-Sectional: Principal Investigators: D. Marcus, R. Buckner, J. Csernansky, J. Morris; P50 AG05681, P01 AG03991, P01 AG026276, R01 AG021910, P20 MH071616, U24 RR021382.

OASIS-2: Longitudinal: Principal Investigators: D. Marcus, R. Buckner, J. Csernansky, J. Morris; P50 AG05681, P01 AG03991, P01 AG026276, R01 AG021910, P20 MH071616, U24 RR021382.

OASIS-3: Longitudinal Multimodal Neuroimaging: 

Principal Investigators: T. Benzinger, D. Marcus, J. Morris; NIH P30 AG066444, P50 AG00561, P30 NS09857781, P01 AG026276, P01 AG003991, R01 AG043434, UL1 TR000448, R01 EB009352. AV-45 doses were provided by Avid Radiopharmaceuticals, a wholly owned subsidiary of Eli Lilly.

OASIS-3\_AV1451: 

Principal Investigators: T. Benzinger, J. Morris; NIH P30 AG066444, AW00006993. AV-1451 doses were provided by Avid Radiopharmaceuticals, a wholly owned subsidiary of Eli Lilly.

OASIS-4: Clinical Cohort: Principal Investigators: T. Benzinger, L. Koenig, P. LaMontagne.

Data used in the preparation of this article were obtained from the MIRIAD database. The MIRIAD investigators did not participate in analysis or writing of this report. The MIRIAD dataset is made available through the support of the UK Alzheimer's Society (Grant RF116). The original data collection was funded through an unrestricted educational grant from GlaxoSmithKline (Grant 6GKC).

Data used in the preparation of this article was obtained on 2023-11-30 from the Parkinson’s Progression Markers Initiative (PPMI) database, RRID:SCR\_006431. For up-to-date information on the study, visit www.ppmi-info.org.

PPMI – a public-private partnership – is funded by the Michael J. Fox Foundation for Parkinson’s Research and funding partners, including 4D Pharma, Abbvie, AcureX, Allergan, Amathus Therapeutics, Aligning Science Across Parkinson's, AskBio, Avid Radiopharmaceuticals, BIAL, BioArctic, Biogen, Biohaven, BioLegend, BlueRock Therapeutics, Bristol-Myers Squibb, Calico Labs, Capsida Biotherapeutics, Celgene, Cerevel Therapeutics, Coave Therapeutics, DaCapo Brainscience, Denali, Edmond J. Safra Foundation, Eli Lilly, Gain Therapeutics, GE HealthCare, Genentech, GSK, Golub Capital, Handl Therapeutics, Insitro, Jazz Pharmaceuticals, Johnson \& Johnson Innovative Medicine, Lundbeck, Merck, Meso Scale Discovery, Mission Therapeutics, Neurocrine Biosciences, Neuron23, Neuropore, Pfizer, Piramal, Prevail Therapeutics, Roche, Sanofi, Servier, Sun Pharma Advanced Research Company, Takeda, Teva, UCB, Vanqua Bio, Verily, Voyager

We also highly appreciate the BraTS and ATLAS organizers for holding the great challenge, and creating the publicly available dataset.

We also highly appreciate BioRender team (https://BioRender.com) for making vivid icons available.

\subsection{Pretraining Dataset}
\label{sec:pretraining_data}
\textbf{UK Biobank (UKBB)}\citep{sudlow2015uk}: The UK Biobank study is a large-scale population-based study that includes a diverse collection of health, demographic, and lifestyle information to explore aging, disease, and cognitive functioning. It provides rich neuroimaging data for studying brain structure, function, and their relationship with various diseases. Our study focused on participants from the imaging-genomics population who had at least one brain MRI scan. UKBB acquired T1-weighted brain MRI data using 3T Siemens Skyra MRI scanner and the Magnetization Prepared Rapid Gradient Echo (MPRAGE) sequence. Some participants (7.79\% in our data freeze) receive a second scan within 2 years after their initial scan. Our collection is based on data available in March, 2024. 

\textbf{Human Connectome Project (HCP)}\citep{harms2018extending, bookheimer2019lifespan}: The Human Connectome Project Lifespan Project focused on gathering data from healthy adults to build a high-quality dataset for comparability with other populations. The HCP-Aging (age 36-60 years) and HCP-Development (age 5-21 years) investigate healthy aging at different development stages. All imaging follow the same Siemens Prisma protocol. Current data release include cross-sectional visit 1 structural brain imaging data.

\textbf{Cambridge Centre for Ageing and Neuroscience (CamCAN)}\citep{taylor2017cambridge}: Cambridge Centre for Ageing and Neuroscience investigates how healthy people retain cognitive abilities as they age. The project uses data from interviews, brain scans, and cognitive experiments. T1-weighted MRI scans are collected using a standardized protocol to ensure high-quality and consistent imaging across the participant cohort.

\textbf{Information eXtraction from Images(IXI)}\citep{rowland2004information}:  Information eXtraction from Images project collected neuroimaging data from normal, healthy subjects with varying scanners at three different hospitals in London. All participants in the IXI study are healthy individuals, offering valuable data for understanding normal aging and brain structure changes, without the confounding effects of neurological conditions.

\subsection{Disease Specific Datasets
}

Multiple neuroimaging datasets aim to identify imaging biomarkers for understanding the trajectory of cognitive decline. They differ in terms of study design, patient selection, patient tracking and image data collection protocol. The diversity of these datasets helps confirm the model's generalizability and reliability for use in diverse healthcare environments, ultimately improving the model’s ability to support early and accurate diagnosis of Alzheimer’s disease.

\textbf{National Alzheimer's Coordinating Center (NACC)}\citep{beekly2007national}: NACC study is a multi-source collection which provides longitudinal data to support research into Alzheimer’s disease, including clinical, genetic, cognitive assessment and imaging data. The project actively follows participants. We include all visits for all participants as of the June 2023 data freeze. The NACC dataset includes MRI scans collected with mixed protocols, including variations in scanner models, acquisition parameters, and protocols. This diversity enables the evaluation of model robustness by testing how well models generalize across different imaging conditions and real-world data variability. 

\textbf{Alzheimer's Disease Neuroimaging Initiative (ADNI)}\citep{petersen2010alzheimer}: Data used in the preparation of this article were obtained from the Alzheimer’s Disease Neuroimaging Initiative (ADNI) database (adni.loni.usc.edu). The ADNI was launched in 2003 as a public-private partnership, led by Principal Investigator Michael W. Weiner, MD. The primary goal of ADNI has been to test whether serial magnetic resonance imaging (MRI), positron emission tomography (PET), other biological markers, and clinical and neuropsychological assessment can be combined to measure the progression of mild cognitive impairment (MCI) and early Alzheimer’s disease (AD).  
We include all available T1 structural MRI images for all participants as of the October 2023 data freeze.

\textbf{OASIS (Open Access Series of Imaging Studies)}\citep{lamontagne2019oasis,marcus2010open, marcus2007open}: The Open Access Series of Imaging Studies (OASIS) dataset primarily targets abnormal aging and Alzheimer’s Disease. It includes a variety of studies with different focuses and includes different participant cohorts. The previously released OASIS1 (cross-sectional) and OASIS2 (longitudinal) datasets have been used for hypothesis-driven analyses, the development of neuroanatomical atlases, and the creation of segmentation algorithms. OASIS-3 is a longitudinal dataset that includes multimodal neuroimaging, clinical, cognitive, and biomarker data focused on normal aging and Alzheimer’s Disease. OASIS-4 consists of MRI, clinical, cognitive, and biomarker data from individuals with memory complaints. Each part is analyzed separately in our study. We include 436 / 373 / 2275 / 181 MRI images (416 / 150 / 1044 / 180 unique subjects) from OASIS1, OASIS2, OASIS3, OASIS4 study.

\textbf{Australian Imaging, Biomarkers \& Lifestyle (AIBL)}\citep{ellis2006australian}: Data used in the preparation of this article was obtained from the Australian Imaging Biomarkers and Lifestyle flagship study of ageing (AIBL). See www.aibl.csiro.au for further details. We include available T1 MRI images from the AIBL study, which are made available via LONI (Laboratory of Neuroimaging, USC) infrastructure.

\textbf{Mild Cognitive Impairment and Alzheimer’s Disease Neuroimaging Database (MIRIAD)}\citep{malone2013miriad}: We include 705 T1 MRI images (of 69 unique subjects) from the MIRIAD study, which focuses on the transition from mild cognitive impairment (MCI) to Alzheimer's disease, including neuroimaging and cognitive data. All scans were conducted on the same 1.5 T Signa MRI scanner (GE Medical systems, Milwaukee, WI) and acquired by the same radiographer. 


\textbf{Parkinson’s Progression Markers Initiative (PPMI)}\citep{marek2011parkinson}: Parkinson’s Progression Markers Initiative dataset investigates biomarkers of Parkinson’s disease through imaging, clinical, and biological data to better understand disease progression. Healthy controls only received imaging as baseline visit, prodromal and Parkinson’s Disease patients might have multiple images. We include data from PPMI participants available in November 2023.

\textbf{OpenNeuro}\citep{schuch2023open}: An MRI dataset of 82 people with epilepsy due to focal cortical dysplasia (FCD) type II and 81 healthy controls was made available in OpenNeuro platform (ds004199). Focal cortical dysplasias (FCDs) are the most common cause of treatment-resistant epilepsy affecting the pediatric population. Most individuals with FCD have seizure onset within the first five years of life and the majority will develop epilepsy by the age of sixteen. We use all T1 MRI scans to develop a model that automatically detects FCD thus identifies epilepsy.

\textbf{EPISURG}\citep{perez2021self}: EPISURG is a clinical dataset of T1-weighted magnetic resonance images (MRI) from 430 epileptic patients who underwent resective brain surgery. All preoperative scans available (n=268) were included in this study as an external validation of model performance.

\textbf{University of California San Francisco Preoperative Diffuse Glioma (UCSF-PDGM)}\citep{calabrese2022university}. The publicly available MRI dataset, consisting of 501 patients with grade 2–4 diffuse gliomas, includes standardized 3-T three-dimensional preoperative MRI protocol neuroimaging data, tumor genetic data, and treatment and survival data. We focus on T1 without and with contrast enhancing(CE) agent, as well as the tumor genetic data for our analysis.

\textbf{Brain Tumor Segmentation (BraTS) Continuous Challenge}\citep{baid2021rsna, menze2014multimodal, bakas2017advancing}: The challenge seeks to identify state-of-the-art segmentation algorithms for brain diffuse glioma patients and their sub-regions. We include T1 MRI scans from BraTS 2023 Adult Glioma challenge training cohort. Each subject received two T1 scans, one with contrast enhancing(CE) agent, one without. All the imaging datasets have been annotated manually by experienced neuro-radiologists. Annotations comprise the enhancing tumor (ET — label 3), the peritumoral edematous/invaded tissue (ED — label 2), and the necrotic tumor core (NCR — label 1). Enhancing tumour refers to regions with visible enhancement on a T1CE sequence after gadolinium administration. Non-enhancing tumour/necrotic tumour core refers to the part of the tumour that does not enhance after gadolinium, typically deep to the enhancement, while oedema/invaded tissue refers to the peritumoral oedematous and/or infiltrated brain parenchyma, typified by hyperintensity on T2 and FLAIR sequences.

\textbf{Anatomical Tracings of Lesions After Stroke(ATLAS R2.0)}\citep{liew2022large}: The goal is to discover automated methods of stroke lesion segmentation in MR images. We include 655 T1 MRI images (of 33 unique subjects) and their manual segmentation masks from the public training collection of the 2022 MICCAI ATLAS challenge. 

This BrainATLAS collection is one of the most diverse and comprehensive public brain MRI datasets, encompassing a wide range of demographics, conditions, and imaging protocols. Its breadth and diversity offer significant potential for advancing AI-driven brain research and enhancing clinical decision-making across various neurological conditions. We provide detailed study population characteristics for each individual dataset in Appendix Table \ref{tab:BrainATLAS_datasets}. A complete list of data usage acknowledgement for can also be found in Appendix Section \ref{sec_app:acknowledgement}. 

\clearpage
\subsection{Study Population in Each Dataset}

We present summary statistics regarding participant demographic and imaging acquisition protocols in Table \ref{tab:BrainATLAS_datasets}. The BrainATLAS represents one of the most diverse and comprehensive collections of publicly available brain MRI data, drawn from 18 studies. Its extensive breadth and variability make it a valuable resource for advancing research in brain imaging and improving the generalizability of AI models in clinical applications.

\begin{table}[h!]
\centering
\begin{tabular}{l|c|c|l|l|l}
\hline
\textbf{Dataset} & \# images & \# subjects & \makecell{Age in yrs\\ (mean ± std)}  & \makecell{Gender\\ (female \%)} & \makecell{Brain volume in mL\\ (mean ± std)} \\ \hline
UKBB & 67294 & 62665 & 64.93 ± 7.69 & 52.81\% & 1156 ± 114 \\ 
HCP & 1350 & 1350 & 38.66 ± 25.76 & 54.97\% & 1179 ± 137 \\ 
CamCAN & 653 & 653 & 54.2 ± 18.5  & 49.35\% & 1156 ± 128 \\ 
IXI & 581 & 581 & 48.65 ± 16.47 & 53.87\% & 1137 ± 131 \\ \hline
NACC & 6956 & 4926 & 71.41 ± 11.25 & 53.34\% & 1034 ± 185 \\ 
ADNI & 6471 & 1817 & 74.68 ± 7.64 & 47.45\% & 1072 ± 116 \\ 
AIBL & 1167 & 664 & 74.56 ± 6.79 & 52.87\% & 1078 ± 110 \\ 
OASIS1 & 436 & 416 & 51.35 ± 25.27 & 61.47\% & 1089 ± 153 \\ 
OASIS2 & 373 & 150 & 77.01 ± 7.64 & 57.10\% & 1045 ± 124 \\ 
OASIS3 & 2275 & 1044 & 70.69 ± 9.18 & 48.39\% & 1061 ± 113 \\ 
OASIS4 & 181 & 180 & 71.09 ± 9.60 & 46.96\% & 1060 ± 116 \\ 
MIRIAD & 705 & 69 & 69.65 ± 6.88 & 55.51\% & 1039 ± 115 \\ 
PPMI & 2032 & 1306 & 63.29 ± 9.58 & 40.59\% & 1119 ± 157 \\ 
OpenNeuro & 163 & 163 & 6.67 ± 2.47 & 46.01\% & 1195 ± 144 \\ 
EPISURG & 268 & 268 & NA & NA & 1102 ± 133 \\ 
UCSF-PDGM & 1980 & 501 & 56.87 ± 15.02 & 40.32\% & NA \\ 
BraTS & 2502 & 1251 & NA & NA & 1103 ± 136 \\ 
ATLAS & 655 & 33 & NA & NA & 1445 ± 124 \\ \hline
\end{tabular}

\begin{minipage}{\textwidth}
    \footnotesize
    \textbf{Footnotes:}
    \begin{itemize}
        \item Excluding missing values and outliers.
        \item Brain volume: BrainSegVol generated by FreeSurfer.
        \item NA: Not available for the analyzed cohort..
    \end{itemize}
\end{minipage}

\caption{Dataset distribution in BrainATLAS. We include a total of 18 distinct neuroimaging dataset from diverse populations.}
\label{tab:BrainATLAS_datasets}
\end{table}

\clearpage
\subsection{Imaging Data Harmonization}
\label{sec_app:image_data_harmonization}

There are a total of 31 steps in the workflow. We conduct a retrospective analysis for using 1500 successful image processing logs to analyze the computational cost for running "recon-all" command in FreeSurfer. We provide a summary of time consumption for different groups of operations in Table \ref{tab:fs_workflow}. On average, the workflow takes 6.11 hours to finish for each 3D image.

1. Motion Correction and Conform

2. NU (Non-Uniform intensity normalization)

3. Talairach transform computation

4. Intensity Normalization 1

5. Skull Strip

6. EM Register (linear volumetric registration)

7. CA Intensity Normalization

8. CA Non-linear Volumetric Registration

9. Remove Neck

10. LTA with Skull

11. CA Label (Volumetric Labeling, ie Aseg) and Statistics

12. Intensity Normalization 2 (start here for control points)

13. White matter segmentation

14. Edit WM With ASeg

15. Fill (start here for wm edits)

16. Tessellation (begins per-hemisphere operations)

17. Smooth1

18. Inflate1

19. QSphere

20. Automatic Topology Fixer

21. Final Surfs (start here for brain edits for pial surf)

22. Smooth2

23. Inflate2

24. Spherical Mapping

25. Spherical Registration

26. Spherical Registration, Contralateral hemisphere

27. Map average curvature to subject

28. Cortical Parcellation - Desikan\_Killiany and Christophe (Labeling)

29. Cortical Parcellation Statistics

30. Cortical Ribbon Mask

31. Cortical Parcellation mapping to Aseg

\begin{table}[ht]
\centering
\begin{tabular}{c|c|p{5cm}|c}
\hline
 & \textbf{Workflow} & \textbf{Explanation} & \makecell{ \textbf{Time} \\\textbf{(mean ± std, hours)}} \\
\hline
\textbf{-autorecon1} & steps 1-5 & Initial processing of the MRI data & 0.36 ± 0.06 \\
\hline
\textbf{-autorecon2} & steps 6-23 & Reconstruction of the brain’s cortical surface and refinement of subcortical structures & 3.78 ± 0.76 \\
\hline
\textbf{-autorecon3} & steps 24-31 & Refinement of segmentation and cortical surfaces & 1.97 ± 0.39 \\
\hline
\end{tabular}
\caption{FreeSurfer main workflow explanation and processing time}
\label{tab:fs_workflow}
\end{table}

\clearpage
\subsection{Brain Anatomical Features}
\label{sec:brain_anatomical_features}

We extract 178 anatomical brain features using FreeSurfer output statistics regarding brain segmentation and surface reconstruction. We summarize the statistical features in Table~\ref{tab:anatomical_features}. These features are categorized into \textit{Volume}, \textit{Surface Area}, and \textit{Thickness}, and collectively capture key structural information of the brain.

\begin{small}
\begin{longtable}{l|l}

\toprule
\textbf{Category} & \textbf{Feature Name} \\
\midrule
\endfirsthead

\toprule
\textbf{Category} & \textbf{Feature Name} \\
\midrule
\endhead

\midrule
\multicolumn{2}{r}{\textit{(continued on next page)}} \\
\endfoot

\bottomrule
\caption{Brain anatomical features categorized by Volume, Surface Area, and Cortex Thickness.}
\label{tab:anatomical_features} \\
\endlastfoot


Volume & Right-VentralDC\_Volume\_mm3 \\
Volume & CC\_Posterior\_Volume\_mm3 \\
Volume & 3rd-Ventricle\_Volume\_mm3 \\
Volume & inferiortemporal\_GrayVol \\
Volume & lateraloccipital\_GrayVol \\
Volume & lateralorbitofrontal\_GrayVol \\
Volume & CC\_Mid\_Posterior\_Volume\_mm3 \\
Volume & paracentral\_GrayVol \\
Volume & VentricleChoroidVol \\
Volume & rostralanteriorcingulate\_GrayVol \\
Volume & middletemporal\_GrayVol \\
Volume & Left-Amygdala\_Volume\_mm3 \\
Volume & TotalGrayVol \\
Volume & CortexVol \\
Volume & fusiform\_GrayVol \\
Volume & precentral\_GrayVol \\
Volume & transversetemporal\_GrayVol \\
Volume & BrainSegVolNotVent \\
Volume & supramarginal\_GrayVol \\
Volume & Left-Putamen\_Volume\_mm3 \\
Volume & parsopercularis\_GrayVol \\
Volume & Left-vessel\_Volume\_mm3 \\
Volume & Brain-Stem\_Volume\_mm3 \\
Volume & Right-Amygdala\_Volume\_mm3 \\
Volume & entorhinal\_GrayVol \\
Volume & 5th-Ventricle\_Volume\_mm3 \\
Volume & BrainSegVol \\
Volume & insula\_GrayVol \\
Volume & Right-Thalamus\_Volume\_mm3 \\
Volume & Right-vessel\_Volume\_mm3 \\
Volume & Right-Inf-Lat-Vent\_Volume\_mm3 \\
Volume & Right-choroid-plexus\_Volume\_mm3 \\
Volume & Left-Hippocampus\_Volume\_mm3 \\
Volume & CC\_Mid\_Anterior\_Volume\_mm3 \\
Volume & caudalanteriorcingulate\_GrayVol \\
Volume & Right-Caudate\_Volume\_mm3 \\
Volume & lingual\_GrayVol \\
Volume & SupraTentorialVolNotVent \\
Volume & Optic-Chiasm\_Volume\_mm3 \\
Volume & Left-Accumbens-area\_Volume\_mm3 \\
Volume & posteriorcingulate\_GrayVol \\
Volume & bankssts\_GrayVol \\
Volume & rhCortexVol \\
Volume & parahippocampal\_GrayVol \\
Volume & Right-Cerebellum-Cortex\_Volume\_mm3 \\
Volume & Left-Thalamus\_Volume\_mm3 \\
Volume & Right-Accumbens-area\_Volume\_mm3 \\
Volume & Left-Cerebellum-Cortex\_Volume\_mm3 \\
Volume & temporalpole\_GrayVol \\
Volume & parstriangularis\_GrayVol \\
Volume & Right-Hippocampus\_Volume\_mm3 \\
Volume & superiorparietal\_GrayVol \\
Volume & 4th-Ventricle\_Volume\_mm3 \\
Volume & pericalcarine\_GrayVol \\
Volume & cuneus\_GrayVol \\
Volume & Left-choroid-plexus\_Volume\_mm3 \\
Volume & superiorfrontal\_GrayVol \\
Volume & Right-Lateral-Ventricle\_Volume\_mm3 \\
Volume & WM-hypointensities\_Volume\_mm3 \\
Volume & rostralmiddlefrontal\_GrayVol \\
Volume & Left-Caudate\_Volume\_mm3 \\
Volume & CerebralWhiteMatterVol \\
Volume & Left-Pallidum\_Volume\_mm3 \\
Volume & Left-Inf-Lat-Vent\_Volume\_mm3 \\
Volume & lhCerebralWhiteMatterVol \\
Volume & CC\_Anterior\_Volume\_mm3 \\
Volume & CC\_Central\_Volume\_mm3 \\
Volume & inferiorparietal\_GrayVol \\
Volume & caudalmiddlefrontal\_GrayVol \\
Volume & CSF\_Volume\_mm3 \\
Volume & Left-Lateral-Ventricle\_Volume\_mm3 \\
Volume & isthmuscingulate\_GrayVol \\
Volume & parsorbitalis\_GrayVol \\
Volume & postcentral\_GrayVol \\
Volume & Right-Cerebellum-White-Matter\_Volume\_mm3 \\
Volume & Left-Cerebellum-White-Matter\_Volume\_mm3 \\
Volume & frontalpole\_GrayVol \\
Volume & superiortemporal\_GrayVol \\
Volume & Right-Putamen\_Volume\_mm3 \\
Volume & precuneus\_GrayVol \\
Volume & Left-VentralDC\_Volume\_mm3 \\
Volume & Right-Pallidum\_Volume\_mm3 \\
Volume & medialorbitofrontal\_GrayVol \\
Volume & Left-WM-hypointensities\_Volume\_mm3 \\
Volume & Right-WM-hypointensities\_Volume\_mm3 \\
Volume & non-WM-hypointensities\_Volume\_mm3 \\
Volume & Left-non-WM-hypointensities\_Volume\_mm3 \\
Volume & Right-non-WM-hypointensities\_Volume\_mm3 \\
Volume & BrainSegVol-to-eTIV \\
\midrule
Surface Area & transversetemporal\_SurfArea \\
Surface Area & temporalpole\_SurfArea \\
Surface Area & supramarginal\_SurfArea \\
Surface Area & rostralmiddlefrontal\_SurfArea \\
Surface Area & posteriorcingulate\_SurfArea \\
Surface Area & parsorbitalis\_SurfArea \\
Surface Area & parsopercularis\_SurfArea \\
Surface Area & middletemporal\_SurfArea \\
Surface Area & lingual\_SurfArea \\
Surface Area & lateralorbitofrontal\_SurfArea \\
Surface Area & caudalmiddlefrontal\_SurfArea \\
Surface Area & caudalanteriorcingulate\_SurfArea \\
Surface Area & bankssts\_SurfArea \\
Surface Area & isthmuscingulate\_SurfArea \\
Surface Area & insula\_SurfArea \\
Surface Area & parahippocampal\_SurfArea \\
Surface Area & parstriangularis\_SurfArea \\
Surface Area & entorhinal\_SurfArea \\
Surface Area & pericalcarine\_SurfArea \\
Surface Area & paracentral\_SurfArea \\
Surface Area & precentral\_SurfArea \\
Surface Area & inferiortemporal\_SurfArea \\
Surface Area & fusiform\_SurfArea \\
Surface Area & cuneus\_SurfArea \\
Surface Area & precuneus\_SurfArea \\
Surface Area & lateraloccipital\_SurfArea \\
Surface Area & supramarginal\_SurfArea \\
Surface Area & superiorfrontal\_SurfArea \\
Surface Area & inferiorparietal\_SurfArea \\
Surface Area & superiorparietal\_SurfArea \\
Surface Area & superiortemporal\_SurfArea \\
Surface Area & rostralanteriorcingulate\_SurfArea \\
Surface Area & superiororbitofrontal\_SurfArea \\
Surface Area & postcentral\_SurfArea \\
Surface Area & medialorbitofrontal\_SurfArea \\
Surface Area & frontalpole\_SurfArea \\
Surface Area & rhCortexArea \\
\midrule
Thickness & insula\_ThickAvg \\
Thickness & transversetemporal\_ThickAvg \\
Thickness & temporalpole\_ThickAvg \\
Thickness & frontalpole\_ThickAvg \\
Thickness & superiortemporal\_ThickAvg \\
Thickness & superiorparietal\_ThickAvg \\
Thickness & superiorfrontal\_ThickAvg \\
Thickness & rostralmiddlefrontal\_ThickAvg \\
Thickness & rostralanteriorcingulate\_ThickAvg \\
Thickness & precuneus\_ThickAvg \\
Thickness & precentral\_ThickAvg \\
Thickness & posteriorcingulate\_ThickAvg \\
Thickness & postcentral\_ThickAvg \\
Thickness & pericalcarine\_ThickAvg \\
Thickness & parstriangularis\_ThickAvg \\
Thickness & parsorbitalis\_ThickAvg \\
Thickness & parsopercularis\_ThickAvg \\
Thickness & paracentral\_ThickAvg \\
Thickness & parahippocampal\_ThickAvg \\
Thickness & middletemporal\_ThickAvg \\
Thickness & medialorbitofrontal\_ThickAvg \\
Thickness & lingual\_ThickAvg \\
Thickness & lateralorbitofrontal\_ThickAvg \\
Thickness & lateraloccipital\_ThickAvg \\
Thickness & isthmuscingulate\_ThickAvg \\
Thickness & inferiortemporal\_ThickAvg \\
Thickness & inferiorparietal\_ThickAvg \\
Thickness & fusiform\_ThickAvg \\
Thickness & entorhinal\_ThickAvg \\
Thickness & cuneus\_ThickAvg \\
Thickness & caudalmiddlefrontal\_ThickAvg \\
Thickness & bankssts\_ThickAvg \\
Thickness & caudalanteriorcingulate\_ThickAvg \\
Thickness & rhCortexThickness \\
Thickness & CortexThickness \\
Thickness & lhCortexThickness \\
Thickness & supramarginal\_CortexThickness \\
Thickness & parahippocampal\_CortexThickness \\
Thickness & lateraloccipital\_CortexThickness \\
Thickness & superiorparietal\_CortexThickness \\
Thickness & precentral\_CortexThickness \\
Thickness & medialorbitofrontal\_CortexThickness \\
Thickness & isthmuscingulate\_CortexThickness \\
Thickness & lingual\_CortexThickness \\
Thickness & rostralmiddlefrontal\_CortexThickness \\
Thickness & entorhinal\_CortexThickness \\
Thickness & cuneus\_CortexThickness \\
Thickness & temporalpole\_CortexThickness \\
Thickness & postcentral\_CortexThickness \\
Thickness & MeanThickness \\
\midrule
Other & SurfaceHoles \\
Other & NumVert \\

\end{longtable}
\end{small}

\subsection{Anatomical Brain Segmentation Workflow}
\label{sec:anatomical_segmentation_workflow}

In this section, we provide more detail into the anatomical brain segmentation workflow used to generate region-level labels from raw MRI volumes. All MRIs included in our pretraining dataset were processed using the FreeSurfer software suite, a well-established tool for structural neuroimaging analysis. FreeSurfer automatically segments the brain into detailed anatomical regions using both volumetric and surface-based techniques.\\
\\
This segmentation provides a rich label space with 112 distinct anatomical regions, which we used as the basis for supervised learning. However, to make model training computationally feasible and clinically meaningful, we applied a post-processing pipeline described in the Methods section to reduce and simplify the label space. The goal was to balance granularity with practicality, preserving essential neuroanatomical information relevant to disease. \\
\\
Appendix Table \ref{tab:brain_region_segmentation} presents the original anatomical regions segmented by FreeSurfer before any preprocessing. Each row corresponds to a distinct label from the raw segmentation. The Region column indicates the numeric label assigned by FreeSurfer, Num MRIs shows how many scans included that region, Percentage Brain Pixels represents the average proportion of brain volume it occupies, and Label provides the anatomical name.\\
\\
Appendix Table \ref{tab:grouped_brain_regions_no_mris} shows the final set of 39 grouped regions used for supervised training. These were obtained by merging symmetrical, related, or minor structures based on clinical relevance and volume thresholds. The Normalized Label column contains the name of each new grouped region, Percentage Brain Pixels indicates the average volume it occupies relative to the whole brain, and Region lists the original FreeSurfer label IDs that were combined. \\

\begin{small}
\begin{longtable}{r|r|r|l}

\toprule
 \textbf{Region} &  \textbf{Number or MRIs} &  \textbf{Percentage Brain Pixels} &                          \textbf{ Label} \\
\midrule
\endfirsthead

\toprule
\textbf{Region} &  \textbf{Number of MRIs} &  \textbf{Percentage Brain Pixels} &                          \textbf{ Label} \\
\midrule
\endhead

\midrule
\multicolumn{4}{r}{\textit{(continued on next page)}} \\
\endfoot

\bottomrule
\caption{Brain region segmentation statistics across all labeled regions before pre-processing.}
\label{tab:brain_region_segmentation}\\
\endlastfoot
      2 &     66709 &               19.67 &      Left-Cerebral-White-Matter \\
      4 &     66709 &                1.22 &          Left-Lateral-Ventricle \\
      5 &     66706 &                0.04 &               Left-Inf-Lat-Vent \\
      7 &     66709 &                1.29 &    Left-Cerebellum-White-Matter \\
      8 &     66709 &                4.64 &          Left-Cerebellum-Cortex \\
     10 &     66709 &                0.62 &            Left-Thalamus-Proper \\
     11 &     66709 &                0.29 &                    Left-Caudate \\
     12 &     66709 &                0.39 &                    Left-Putamen \\
     13 &     66709 &                0.17 &                   Left-Pallidum \\
     14 &     66709 &                0.12 &                   3rd-Ventricle \\
     15 &     66709 &                0.15 &                   4th-Ventricle \\
     16 &     66709 &                1.88 &                      Brain-Stem \\
     17 &     66709 &                0.35 &                Left-Hippocampus \\
     18 &     66709 &                0.13 &                   Left-Amygdala \\
     24 &     66709 &                0.10 &                             CSF \\
     26 &     66709 &                0.04 &             Left-Accumbens-area \\
     28 &     66709 &                0.35 &                  Left-VentralDC \\
     30 &     64102 &                0.00 &                     Left-vessel \\
     31 &     66709 &                0.07 &             Left-choroid-plexus \\
     41 &     66709 &               19.75 &     Right-Cerebral-White-Matter \\
     43 &     66709 &                1.11 &         Right-Lateral-Ventricle \\
     44 &     66708 &                0.04 &              Right-Inf-Lat-Vent \\
     46 &     66709 &                1.23 &   Right-Cerebellum-White-Matter \\
     47 &     66709 &                4.79 &         Right-Cerebellum-Cortex \\
     49 &     66709 &                0.62 &           Right-Thalamus-Proper \\
     50 &     66709 &                0.30 &                   Right-Caudate \\
     51 &     66709 &                0.40 &                   Right-Putamen \\
     52 &     66709 &                0.17 &                  Right-Pallidum \\
     53 &     66709 &                0.37 &               Right-Hippocampus \\
     54 &     66709 &                0.15 &                  Right-Amygdala \\
     58 &     66709 &                0.05 &            Right-Accumbens-area \\
     60 &     66709 &                0.35 &                 Right-VentralDC \\
     62 &     63318 &                0.00 &                    Right-vessel \\
     63 &     66709 &                0.07 &            Right-choroid-plexus \\
     77 &     66709 &                0.21 &              WM-hypointensities \\
     85 &     66707 &                0.02 &                    Optic-Chiasm \\
    251 &     66709 &                0.10 &                    CC\_Posterior \\
    252 &     66709 &                0.05 &                CC\_Mid\_Posterior \\
    253 &     66709 &                0.05 &                      CC\_Central \\
    254 &     66709 &                0.05 &                 CC\_Mid\_Anterior \\
    255 &     66709 &                0.08 &                     CC\_Anterior \\
   1001 &     66709 &                0.18 &                 ctx-lh-bankssts \\
   1002 &     66708 &                0.14 &  ctx-lh-caudalanteriorcingulate \\
   1003 &     66709 &                0.51 &      ctx-lh-caudalmiddlefrontal \\
   1005 &     66709 &                0.26 &                   ctx-lh-cuneus \\
   1006 &     66706 &                0.17 &               ctx-lh-entorhinal \\
   1007 &     66709 &                0.79 &                 ctx-lh-fusiform \\
   1008 &     66709 &                0.94 &         ctx-lh-inferiorparietal \\
   1009 &     66709 &                0.91 &         ctx-lh-inferiortemporal \\
   1010 &     66709 &                0.22 &         ctx-lh-isthmuscingulate \\
   1011 &     66709 &                1.02 &         ctx-lh-lateraloccipital \\
   1012 &     66709 &                0.62 &     ctx-lh-lateralorbitofrontal \\
   1013 &     66709 &                0.54 &                  ctx-lh-lingual \\
   1014 &     66708 &                0.44 &      ctx-lh-medialorbitofrontal \\
   1015 &     66709 &                0.88 &           ctx-lh-middletemporal \\
   1016 &     66705 &                0.18 &          ctx-lh-parahippocampal \\
   1017 &     66708 &                0.29 &              ctx-lh-paracentral \\
   1018 &     66709 &                0.37 &          ctx-lh-parsopercularis \\
   1019 &     66709 &                0.19 &            ctx-lh-parsorbitalis \\
   1020 &     66709 &                0.28 &         ctx-lh-parstriangularis \\
   1021 &     66709 &                0.16 &            ctx-lh-pericalcarine \\
   1022 &     66709 &                0.78 &              ctx-lh-postcentral \\
   1023 &     66709 &                0.25 &       ctx-lh-posteriorcingulate \\
   1024 &     66709 &                1.10 &               ctx-lh-precentral \\
   1025 &     66709 &                0.78 &                ctx-lh-precuneus \\
   1026 &     66707 &                0.22 & ctx-lh-rostralanteriorcingulate \\
   1027 &     66709 &                1.20 &     ctx-lh-rostralmiddlefrontal \\
   1028 &     66709 &                1.84 &          ctx-lh-superiorfrontal \\
   1029 &     66709 &                1.08 &         ctx-lh-superiorparietal \\
   1030 &     66709 &                1.00 &         ctx-lh-superiortemporal \\
   1031 &     66709 &                0.91 &            ctx-lh-supramarginal \\
   1032 &     66709 &                0.08 &              ctx-lh-frontalpole \\
   1033 &     66709 &                0.22 &             ctx-lh-temporalpole \\
   1034 &     66709 &                0.10 &       ctx-lh-transversetemporal \\
   1035 &     66709 &                0.57 &                   ctx-lh-insula \\
   2001 &     66709 &                0.17 &                 ctx-rh-bankssts \\
   2002 &     66708 &                0.16 &  ctx-rh-caudalanteriorcingulate \\
   2003 &     66709 &                0.48 &      ctx-rh-caudalmiddlefrontal \\
   2005 &     66709 &                0.27 &                   ctx-rh-cuneus \\
   2006 &     66709 &                0.16 &               ctx-rh-entorhinal \\
   2007 &     66709 &                0.78 &                 ctx-rh-fusiform \\
   2008 &     66709 &                1.16 &         ctx-rh-inferiorparietal \\
   2009 &     66709 &                0.90 &         ctx-rh-inferiortemporal \\
   2010 &     66707 &                0.20 &         ctx-rh-isthmuscingulate \\
   2011 &     66709 &                1.07 &         ctx-rh-lateraloccipital \\
   2012 &     66709 &                0.61 &     ctx-rh-lateralorbitofrontal \\
   2013 &     66708 &                0.57 &                  ctx-rh-lingual \\
   2014 &     66709 &                0.47 &      ctx-rh-medialorbitofrontal \\
   2015 &     66709 &                0.98 &           ctx-rh-middletemporal \\
   2016 &     66706 &                0.16 &          ctx-rh-parahippocampal \\
   2017 &     66709 &                0.32 &              ctx-rh-paracentral \\
   2018 &     66709 &                0.31 &          ctx-rh-parsopercularis \\
   2019 &     66709 &                0.23 &            ctx-rh-parsorbitalis \\
   2020 &     66709 &                0.33 &         ctx-rh-parstriangularis \\
   2021 &     66707 &                0.18 &            ctx-rh-pericalcarine \\
   2022 &     66709 &                0.75 &              ctx-rh-postcentral \\
   2023 &     66709 &                0.25 &       ctx-rh-posteriorcingulate \\
   2024 &     66709 &                1.07 &               ctx-rh-precentral \\
   2025 &     66709 &                0.80 &                ctx-rh-precuneus \\
   2026 &     66709 &                0.16 & ctx-rh-rostralanteriorcingulate \\
   2027 &     66709 &                1.24 &     ctx-rh-rostralmiddlefrontal \\
   2028 &     66709 &                1.76 &          ctx-rh-superiorfrontal \\
   2029 &     66709 &                1.05 &         ctx-rh-superiorparietal \\
   2030 &     66709 &                0.95 &         ctx-rh-superiortemporal \\
   2031 &     66709 &                0.82 &            ctx-rh-supramarginal \\
   2032 &     66709 &                0.10 &              ctx-rh-frontalpole \\
   2033 &     66709 &                0.23 &             ctx-rh-temporalpole \\
   2034 &     66709 &                0.07 &       ctx-rh-transversetemporal \\
   2035 &     66709 &                0.56 &                   ctx-rh-insula \\
     80 &      2062 &                0.00 &          non-WM-hypointensities \\
     72 &       828 &                0.00 &                   5th-Ventricle \\
     29 &        59 &                0.00 &               Left-undetermined \\
\end{longtable}
\end{small}

\begin{small}
\begin{longtable}{l|r|l}

\toprule
            \textbf{Normalized Label} &  \textbf{Percentage Brain Pixels} &                    \textbf{Region} \\
\midrule
\endfirsthead

\toprule
            \textbf{Normalized Label} &  \textbf{Percentage Brain Pixels} &                    \textbf{Region} \\
\midrule
\endhead
\midrule
\multicolumn{3}{r}{\textit{(continued on next page)}} \\
\endfoot
\bottomrule
\caption{Grouped brain region segmentation statistics after pre-processing.}
\label{tab:grouped_brain_regions_no_mris}
\endlastfoot
              Accumbens-area &                0.08 &                  [26, 58] \\
                    Amygdala &                0.28 &                  [18, 54] \\
                  Brain-Stem &                1.88 &                      [16] \\
                          CC &                0.33 & [251, 252, 253, 254, 255] \\
                     Caudate &                0.59 &                  [11, 50] \\
           Cerebellum-Cortex &                9.43 &                   [8, 47] \\
     Cerebellum-White-Matter &                2.52 &                   [7, 46] \\
       Cerebral-White-Matter &               39.43 &                   [2, 41] \\
                 Hippocampus &                0.72 &                  [17, 53] \\
           Lateral-Ventricle &                2.33 &                   [4, 43] \\
                    Pallidum &                0.34 &                  [13, 52] \\
                     Putamen &                0.79 &                  [12, 51] \\
             Thalamus-Proper &                1.23 &                  [10, 49] \\
          WM-hypointensities &                0.21 &                      [77] \\
              ctx-entorhinal &                0.33 &              [1006, 2006] \\
                ctx-fusiform &                1.58 &              [1007, 2007] \\
        ctx-inferiorparietal &                2.10 &              [1008, 2008] \\
        ctx-inferiortemporal &                1.81 &              [1009, 2009] \\
                  ctx-insula &                1.13 &              [1035, 2035] \\
        ctx-lateraloccipital &                2.09 &              [1011, 2011] \\
    ctx-lateralorbitofrontal &                1.22 &              [1012, 2012] \\
                 ctx-lingual &                1.12 &              [1013, 2013] \\
     ctx-medialorbitofrontal &                0.90 &              [1014, 2014] \\
          ctx-middletemporal &                1.86 &              [1015, 2015] \\
         ctx-parahippocampal &                0.34 &              [1016, 2016] \\
         ctx-parsopercularis &                0.67 &              [1018, 2018] \\
           ctx-parsorbitalis &                0.42 &              [1019, 2019] \\
        ctx-parstriangularis &                0.62 &              [1020, 2020] \\
             ctx-postcentral &                1.53 &              [1022, 2022] \\
      ctx-posteriorcingulate &                0.50 &              [1023, 2023] \\
              ctx-precentral &                2.17 &              [1024, 2024] \\
               ctx-precuneus &                1.58 &              [1025, 2025] \\
ctx-rostralanteriorcingulate &                0.37 &              [1026, 2026] \\
    ctx-rostralmiddlefrontal &                2.44 &              [1027, 2027] \\
         ctx-superiorfrontal &                3.60 &              [1028, 2028] \\
        ctx-superiorparietal &                2.13 &              [1029, 2029] \\
        ctx-superiortemporal &                1.94 &              [1030, 2030] \\
           ctx-supramarginal &                1.73 &              [1031, 2031] \\
            ctx-temporalpole &                0.45 &              [1033, 2033] \\
\end{longtable}
\end{small}

\section{Preliminary Analyses}
\label{sec:preliminary_analyses}

In this section of the Appendix, we report the preliminary analyses used to select the anatomical model, label encoding, and supervised baseline architecture. For all preliminary analyses, model selection was performed exclusively using a validation subset comprising 10\% of the ADNI training split. Results on external cohorts are also reported for completeness, but were not used for model selection. Validation results used for model selection are underlined in the corresponding tables.

\subsection{Preliminary Analysis of Linear Model and Nonlinear Model Using Anatomical Features}
\label{sec:lr_vs_rf}

We extracted 178 FreeSurfer-derived anatomical features from the \texttt{recon-all} pipeline, spanning regional brain volumes, cortical surface areas, and cortical thickness values. The complete list of features is reported in Table~\ref{tab:anatomical_features}. To verify whether more complex tabular models offered an advantage over logistic regression, we performed an additional comparison between linear and non-linear models using dementia detection as an example task.

Hyperparameters were selected using 5-fold cross-validation. For the linear model, we evaluated both no regularization and L2 regularization with coefficient 1, consistent with the default setting in scikit-learn logistic regression. For the non-linear model, we used a random forest with \texttt{n\_estimators} in \{10, 50, 100\}, \texttt{max\_depth} in \{5, 10, 20\}, and \texttt{criterion} in \{\texttt{gini}, \texttt{entropy}\}. To assess whether the comparison depended on sample size, we repeated the experiment using 10\%, 30\%, and 100\% of the training data. As shown in Table~\ref{tab:lr_vs_rf_dementia}, performance differences were generally small, which supports the use of logistic regression as the main anatomical baseline.

\begin{table}[!ht]
\centering

\setlength{\tabcolsep}{4pt}
\renewcommand{\arraystretch}{1.15}

\begin{tabular}{l|cc|cc|cc}
\toprule
Dataset &
Linear 10\% & Non-Linear 10\% &
Linear 30\% & Non-Linear 30\% &
Linear 100\% & Non-Linear 100\% \\
\midrule

ADNI &
\makecell{0.6882 \\ {[0.6566, 0.7159]}} &
\makecell{0.7055 \\ {[0.6719, 0.7395]}} &
\makecell{0.7470 \\ {[0.7165, 0.7798]}} &
\makecell{0.7398 \\ {[0.7144, 0.7721]}} &
\makecell{0.7586 \\ {[0.7279, 0.7874]}} &
\makecell{0.7529 \\ {[0.7203, 0.7834]}} \\

NACC &
\makecell{0.7221 \\ {[0.6723, 0.7693]}} &
\makecell{0.7510 \\ {[0.7187, 0.7901]}} &
\makecell{0.7931 \\ {[0.7527, 0.8463]}} &
\makecell{0.8061 \\ {[0.7546, 0.8516]}} &
\makecell{0.7954 \\ {[0.7566, 0.8531]}} &
\makecell{0.8216 \\ {[0.7741, 0.8661]}} \\

AIBL &
\makecell{0.7111 \\ {[0.6710, 0.7526]}} &
\makecell{0.7582 \\ {[0.7162, 0.8010]}} &
\makecell{0.8004 \\ {[0.7576, 0.8468]}} &
\makecell{0.7933 \\ {[0.7421, 0.8424]}} &
\makecell{0.8105 \\ {[0.7682, 0.8522]}} &
\makecell{0.8261 \\ {[0.7847, 0.8651]}} \\

OASIS1 &
\makecell{0.8328 \\ {[0.7867, 0.8751]}} &
\makecell{0.7967 \\ {[0.7421, 0.8356]}} &
\makecell{0.8168 \\ {[0.7673, 0.8538]}} &
\makecell{0.8150 \\ {[0.7637, 0.8671]}} &
\makecell{0.8066 \\ {[0.7590, 0.8486]}} &
\makecell{0.8245 \\ {[0.7706, 0.8772]}} \\

OASIS2 &
\makecell{0.6796 \\ {[0.6234, 0.7394]}} &
\makecell{0.6825 \\ {[0.6219, 0.7313]}} &
\makecell{0.7085 \\ {[0.6393, 0.7522]}} &
\makecell{0.7038 \\ {[0.6429, 0.7598]}} &
\makecell{0.7378 \\ {[0.6778, 0.7867]}} &
\makecell{0.7710 \\ {[0.7180, 0.8173]}} \\

OASIS3 &
\makecell{0.6621 \\ {[0.5885, 0.7348]}} &
\makecell{0.6847 \\ {[0.6018, 0.7619]}} &
\makecell{0.7124 \\ {[0.6396, 0.7831]}} &
\makecell{0.7009 \\ {[0.6135, 0.7890]}} &
\makecell{0.7258 \\ {[0.6488, 0.7949]}} &
\makecell{0.7192 \\ {[0.6310, 0.7955]}} \\

OASIS4 &
\makecell{0.6326 \\ {[0.5346, 0.7616]}} &
\makecell{0.6291 \\ {[0.5173, 0.7602]}} &
\makecell{0.6301 \\ {[0.5293, 0.7618]}} &
\makecell{0.6376 \\ {[0.5155, 0.7541]}} &
\makecell{0.6250 \\ {[0.5130, 0.7517]}} &
\makecell{0.6330 \\ {[0.5103, 0.7514]}} \\

MIRIAD &
\makecell{0.9708 \\ {[0.9584, 0.9825]}} &
\makecell{0.9799 \\ {[0.9678, 0.9890]}} &
\makecell{0.9683 \\ {[0.9542, 0.9806]}} &
\makecell{0.9771 \\ {[0.9687, 0.9856]}} &
\makecell{0.9733 \\ {[0.9602, 0.9847]}} &
\makecell{0.9855 \\ {[0.9784, 0.9912]}} \\

\bottomrule
\end{tabular}
\caption{Model comparison on dementia detection task with anatomical features. External cohort performance (95\% CI) per dataset is also included.}
\label{tab:lr_vs_rf_dementia}
\end{table}

\FloatBarrier

\subsection{Preliminary Analysis of Label Encoding}
\label{sec:label_encoding}

Because dementia progression is not a strictly discrete process, we evaluated different label encodings for the dementia task. In the \emph{soft} encoding, labels were assigned as CN $=0$, MCI $=0.5$, and AD $=1$, and models were trained using log loss. In the \emph{rigid} encoding, both MCI and AD were treated as positive cases, giving CN $=0$ and MCI/AD $=1$. We also evaluated a standard three-class encoding with labels CN $=0$, MCI $=1$, and AD $=2$.

Preliminary experiments with a linear model using anatomical features showed that the soft-label formulation achieved the best performance on ADNI as shown in (Table~\ref{tab:label_encoding_performance}). For this reason, we adopted the soft-label formulation in the main dementia experiments.

\begin{table}[ht]
\centering
\begin{tabular}{c|c|c|c|c|c}
\hline
\textbf{Dataset} & \textbf{Label Encoding} & \textbf{AD vs CN} & \textbf{AD vs MCI} & \textbf{MCI vs CN} & \textbf{MACRO} \\
\hline
ADNI & Soft &
\makecell{0.8873\\{[0.8551, 0.9201]}} &
\makecell{0.7445\\{[0.7099, 0.7742]}} &
\makecell{0.6926\\{[0.6532, 0.7389]}} &
\makecell{0.7586\\{[0.7279, 0.7874]}}\\
ADNI & Rigid &
\makecell{0.8442\\{[0.8165, 0.8762]}} &
\makecell{0.7184\\{[0.6887, 0.7412]}} &
\makecell{0.6565\\{[0.6092, 0.6997]}} &
\makecell{0.7397\\{[0.7088, 0.7703]}}\\
ADNI & Multi-class &
\makecell{0.7107\\{[0.6786, 0.7395]}} &
\makecell{0.6284\\{[0.5972, 0.6586]}} &
\makecell{0.5927\\{[0.5503, 0.6425]}} &
\makecell{0.6439\\{[0.6134, 0.6736]}}\\
\hline
NACC & Soft &
\makecell{0.9287\\{[0.8977, 0.9532]}} &
\makecell{0.8588\\{[0.8239, 0.8913]}} &
\makecell{0.7850\\{[0.7421, 0.8231]}} &
\makecell{0.8575\\{[0.8301, 0.8842]}}\\
NACC & Rigid &
\makecell{0.9157\\{[0.8853, 0.9419]}} &
\makecell{0.7870\\{[0.7538, 0.8140]}} &
\makecell{0.7085\\{[0.6719, 0.7385]}} &
\makecell{0.8037\\{[0.7696, 0.8323]}}\\
NACC & Multi-class &
\makecell{0.8215\\{[0.7913, 0.8563]}} &
\makecell{0.6824\\{[0.6499, 0.7102]}} &
\makecell{0.6596\\{[0.6172, 0.7014]}} &
\makecell{0.7212\\{[0.6941, 0.7541]}}\\
\hline
\end{tabular}
\caption{Comparison of label encoding methods with a linear model using brain anatomical features.}
\label{tab:label_encoding_performance}
\end{table}

\subsection{Preliminary Analysis of Baseline Neural Network Model Choice}
\label{sec:preliminary_analysis}

We performed a preliminary analysis to select a reasonably strong supervised CNN baseline for comparison with the transformer-based models. The analysis was carried out on the ADNI dementia detection task and compared ResNet architectures with different input sizes. Specifically, we evaluated ResNet-18 with input sizes of $(128,128,128)$ and $(160,160,160)$, and ResNet-50 with input size $(128,128,128)$.

As shown in Table~\ref{tab:baseline_model_choice}, ResNet-18 with input size $(128,128,128)$ achieved the best overall performance and was therefore adopted as the CNN baseline in the main experiments.

\begin{table}[h]
\centering
\begin{tabular}{c|c|c|c|c|c}
\hline
\textbf{Model Type} & \textbf{Input size} & \textbf{AD vs CN} & \textbf{AD vs MCI} & \textbf{MCI vs CN} & \textbf{MACRO} \\
\hline
\textbf{ResNet 18} & (128, 128, 128) &
\makecell{0.8674\\{[0.8328, 0.9011]}} &
\makecell{0.7268\\{[0.6885, 0.7734]}} &
\makecell{0.6852\\{[0.6449, 0.7286]}} &
\makecell{0.7598\\{[0.7277, 0.7889]}} \\
\textbf{ResNet 18} & (160, 160, 160) &
\makecell{0.8106\\{[0.7792, 0.8421]}} &
\makecell{0.6737\\{[0.6352, 0.7178]}} &
\makecell{0.6550\\{[0.6082, 0.7002]}} &
\makecell{0.7141\\{[0.6839, 0.7466]}}\\
\textbf{ResNet 50} & (128, 128, 128) &
\makecell{0.8382\\{[0.8014, 0.8671]}} &
\makecell{0.7029\\{[0.6711, 0.7401]}} &
\makecell{0.6638\\{[0.6229, 0.7073]}} &
\makecell{0.7350\\{[0.7012, 0.7653]}} \\
\hline
\end{tabular}
\caption{Comparison of model performance with different input sizes and model complexity. We adopt the best performing model as the baseline supervised model.}
\label{tab:baseline_model_choice}
\end{table}

\section{Additional Model and Training Details}
\label{sec:additional_model_training_details}

\subsection{Additional details on MAE pretraining and downstream adaptation}
\label{sec:mae_additional}
The pretraining cohort consisted of UKBB, HCP, CamCAN, and IXI. Ten percent of UKBB participants were excluded from MAE and ASP optimization and used for linear-probe monitoring during pretraining. This participant-level holdout, comprising 6,950 MRI scans, was subsequently used as the internal held-out cohort for age estimation.

Following pretraining, the encoder was adapted using the task-specific strategies described in the main Methods: linear probing for Parkinson's disease and IDH mutation prediction, and LoRA for the remaining downstream tasks. Full fine-tuning was evaluated as an ablation against LoRA. For regression tasks, we used a one-dimensional output and optimized the model with mean squared error (MSE) loss. For classification tasks, we used a two-dimensional output and optimized with binary cross-entropy.

For healthy aging, we finetuned the model on healthy subjects from the pretraining cohort and evaluated generalization on healthy subjects from ADNI and NACC. For dementia diagnosis, we finetuned on ADNI and evaluated on a held-out ADNI test cohort as well as external cohorts from NACC, AIBL, OASIS1-4, and MIRIAD. For Parkinson’s disease diagnosis, tumor IDH mutation detection, and epilepsy diagnosis, we used task-specific train/test splits from PPMI, UCSF-PDGM, and OpenNeuro, respectively.

\subsection{Additional details on ASP architecture}
\label{sec:seg_architecture}

The ASP decoder was designed to make the ViT embedding carry anatomical information, rather than allowing a large segmentation head to solve the task on its own. For this reason, we avoid skip connections and use a small bottleneck decoder that receives the global image embedding as input. This differs from standard U-Net-style segmentation models, where local features from early layers are repeatedly fused into the decoder.

To handle multiple anatomical regions efficiently, the decoder is conditioned on a learned region embedding. At each training step, one anatomical region is selected and the model predicts a binary mask for that region. This avoids producing dense logits for all 39 regions at once, which would be memory intensive for full 3D volumes.

The decoder starts from a 768-dimensional image embedding summed with the selected region embedding. This representation is then passed through six 3D transposed-convolution blocks, which upsample the feature map from \(1^3\) to \(128^3\). Each block is followed by BatchNorm and ReLU, except for the final layer. The full decoder contains only around 15 million parameters, which is much smaller than a typical 3D U-Net. This keeps the segmentation head lightweight and encourages the ViT encoder to learn the anatomical semantics instead of delegating the task to the decoder.

\begin{figure}[H]
  \centering
  \includegraphics[width=0.8\textwidth]{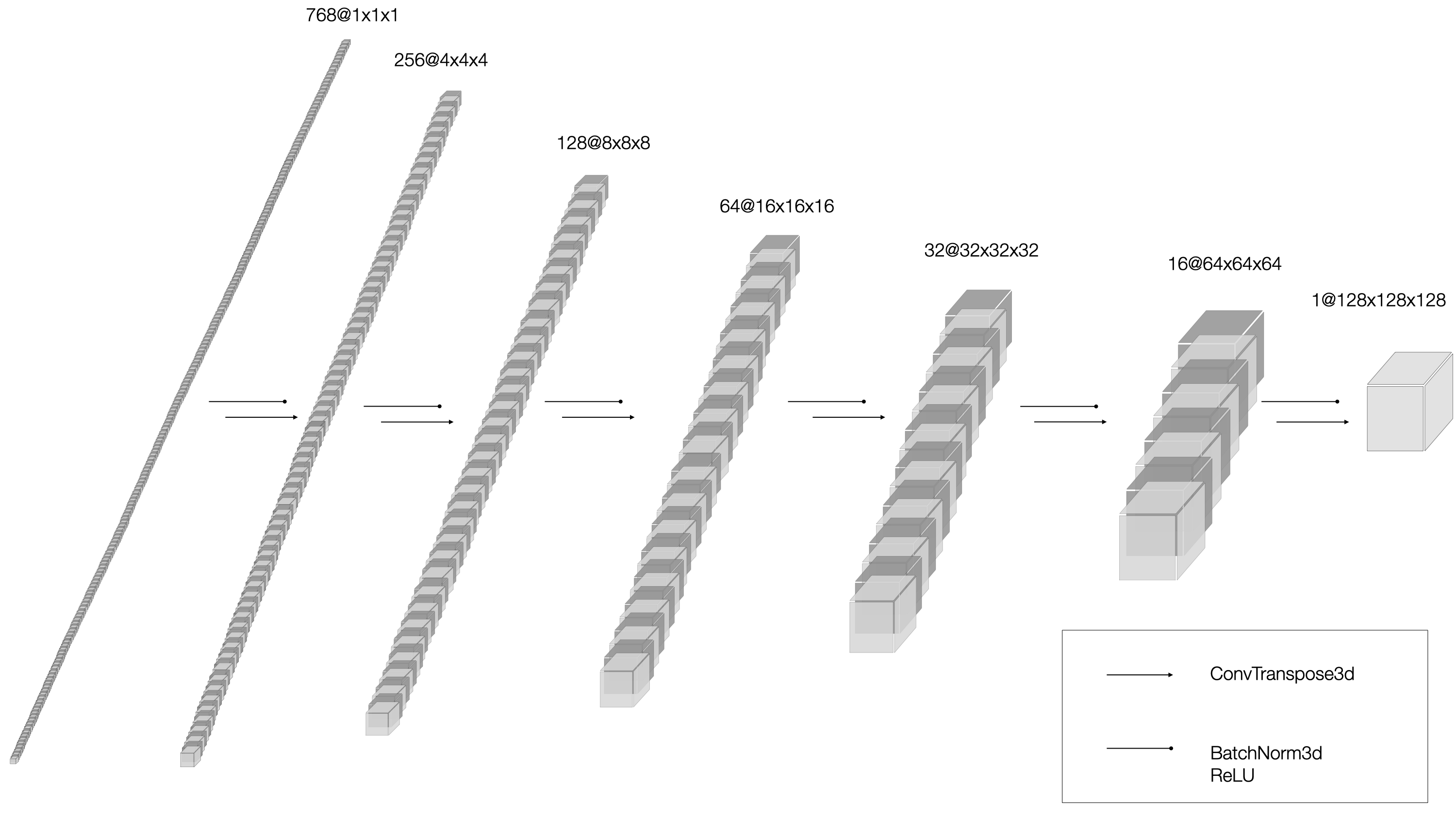}
  \caption{Lightweight region-conditioned 3D mask decoder used for ASP.}
  \label{fig:seg_decoder}
\end{figure}

\clearpage
\subsection{Ablation Study on ASP Training}
\label{ablation_study}

We performed an ablation study to evaluate the main training choices used in ASP. For Alzheimer’s disease and epilepsy detection, we compared full fine-tuning (FF) with LoRA. As shown in Table~\ref{tab:sc_LoRA_classification}, both methods performed similarly for Alzheimer’s disease, while LoRA improved epilepsy detection. For age estimation, we additionally evaluated several ASP training ablations, including removing MAE alternation, encoder freezing, MAE-only pretraining, and training from scratch. The results in Table~\ref{tab:healthy_aging_results} correspond to the variants shown in Figure~\ref{fig:ASP}(e). We evaluate age estimation on UK Biobank as the internal test cohort and NACC as the external test cohort. To avoid confusion with masked autoencoding, we report mean absolute error as ``absolute error'' in the table.

The full ASP model achieves the best performance on both cohorts. Removing alternating masked-autoencoder reconstruction updates reduces performance, and removing both alternating updates and the initial encoder-freezing stage reduces it further. The masked-autoencoder-only model remains a strong baseline, but it is outperformed by the ASP variants that add anatomical supervision. Training from scratch performs worst, showing that ASP benefits from starting from a pretrained masked autoencoder.

\begin{table}[h]
\centering
\small
\begin{tabular}{cccc}
\toprule
\textbf{Task} & \shortstack[c]{\textbf{Evaluation}\\\textbf{Setting}} & \textbf{ASP (FF)} & \textbf{ASP (LoRA)} \\
\midrule


\multirow{8}{*}{\shortstack[c]{AD\\(Train: ADNI)}}
& ADNI   & 0.7800 & \textbf{0.7815} \\
& NACC   & 0.8260 & \textbf{0.8264} \\
& AIBL   & \textbf{0.8318} & 0.8232 \\
& OASIS1 & \textbf{0.8733} & 0.8083 \\
& OASIS2 & 0.8092 & \textbf{0.8439} \\
& OASIS3 & \textbf{0.7454} & 0.7376 \\
& OASIS4 & \textbf{0.6622} & 0.6457 \\
& MIRIAD & 0.9815 & \textbf{0.9866} \\
\midrule

\multirow{8}{*}{\shortstack[c]{AD\\(Train: NACC)}}
& ADNI   & \textbf{0.7676} & 0.7665 \\
& NACC   & 0.8438 & \textbf{0.8570} \\
& AIBL   & \textbf{0.8314} & 0.8216 \\
& OASIS1 & 0.9091 & \textbf{0.9120} \\
& OASIS2 & 0.8105 & \textbf{0.8308} \\
& OASIS3 & \textbf{0.7748} & 0.7658 \\
& OASIS4 & \textbf{0.6658} & 0.6502 \\
& MIRIAD & \textbf{0.9942} & 0.9914 \\
\midrule

\multirow{2}{*}{\shortstack[c]{Epilepsy Detection}}
& Internal & 0.6398 & \textbf{0.7507} \\
& External & 0.8359 & \textbf{0.8716} \\
\bottomrule
\end{tabular}
\caption{Comparison of ASP (FF/Full-Finetuning) and ASP (LoRA) AUROC performance across classification tasks.}
\label{tab:sc_LoRA_classification}
\end{table}

\begin{table*}[h!]
\centering
\scriptsize
\setlength{\tabcolsep}{4pt}
\renewcommand{\arraystretch}{1.15}
\resizebox{0.92\textwidth}{!}{%
\begin{tabular}{lccc ccc}
\toprule
\multirow{2}{*}{\textbf{Model}} 
& \multicolumn{3}{c}{\textbf{Internal: UK Biobank}} 
& \multicolumn{3}{c}{\textbf{External: NACC}} \\
\cmidrule(lr){2-4} \cmidrule(lr){5-7}
& \textbf{Pearson $r \uparrow$} 
& \textbf{MSE $\downarrow$} 
& \textbf{Abs. error $\downarrow$}
& \textbf{Pearson $r \uparrow$} 
& \textbf{MSE $\downarrow$} 
& \textbf{Abs. error $\downarrow$} \\
\midrule

\textbf{ASP (LoRA)}
& \textbf{0.9361} & \textbf{10.38} & \textbf{2.54}
& \textbf{0.8589} & \textbf{29.26} & \textbf{4.19} \\

ASP (Full-Finetuning)
& 0.9333 & 10.83 & 2.60
& 0.8464 & 32.06 & 4.32 \\

No masked-autoencoder alternation
& 0.9291 & 11.49 & 2.67
& 0.8324 & 34.51 & 4.46 \\

No alternation and no freezing
& 0.9220 & 12.59 & 2.78
& 0.7987 & 40.50 & 4.71 \\

Masked autoencoder only
& 0.9159 & 13.61 & 2.89
& 0.8043 & 45.35 & 5.20 \\

From scratch
& 0.8736 & 19.88 & 3.51
& 0.7407 & 55.84 & 5.79 \\

\bottomrule
\end{tabular}%
}
\caption{ASP training ablation evaluated on age estimation. Absolute error denotes mean absolute error in years.}
\label{tab:healthy_aging_results}
\end{table*}

\clearpage




\end{appendices}


\end{document}